\RequirePackage{fix-cm}
\documentclass[twocolumn]{svjour3}          
\smartqed  
\usepackage{graphicx}
\usepackage[sort]{natbib}
\usepackage{hyperref}

\usepackage{amssymb}
\usepackage{amsmath}
\usepackage{amsfonts}
\usepackage{rotating,siunitx} %
\usepackage{tabularx}
\usepackage{hyphenat}
\usepackage{longtable}
\usepackage{algorithm}
\usepackage{algorithmic}
\usepackage{natbib}
\usepackage{bibunits}
\usepackage[labelformat=simple]{subfig}
\usepackage{soul}
\usepackage{xcolor}

\begin{document}

\title{Curriculum Learning: A Survey\thanks{This work was supported by a grant of the Romanian Ministry of Education and Research, CNCS - UEFISCDI, project number PN-III-P1-1.1-TE-2019-0235, within PNCDI III. This article has also benefited from the support of the Romanian Young Academy, which is funded by Stiftung Mercator and the Alexander von Humboldt Foundation for the period 2020-2022. This work was also supported by European Union's Horizon 2020 research and innovation programme under grant number 951911 - AI4Media.}
}

\author{Petru Soviany         \and
        Radu Tudor Ionescu \and
        Paolo Rota \and
        Nicu Sebe
}


\institute{P. Soviany \at
              Department of Computer Science,
              University of Bucharest,
              Bucharest 010014, Romania
           \and
           R.T. Ionescu \at
              Department of Computer Science and Romanian Young Academy,
              University of Bucharest,
              Bucharest 010014, Romania  \\
              \email{raducu.ionescu@gmail.com}
           \and
           P. Rota \at
              Department of Information Engineering and Computer Science,
            University of Trento,
            Povo-Trento 38123, Italy
            \and
           N. Sebe \at
              Department of Information Engineering and Computer Science,
            University of Trento,
            Povo-Trento 38123, Italy
}

\date{Received: date / Accepted: date}

\maketitle

\begin{abstract}
Training machine learning models in a meaningful order, from the easy samples to the hard ones, using curriculum learning can provide performance improvements over the standard training approach based on random data shuffling, without any additional computational costs. Curriculum learning strategies have been successfully employed in all areas of machine learning, in a wide range of tasks. However, the necessity of finding a way to rank the samples from easy to hard, as well as the right pacing function for introducing more difficult data can limit the usage of the curriculum approaches. In this survey, we show how these limits have been tackled in the literature, and we present different curriculum learning instantiations for various tasks in machine learning. We construct a multi-perspective taxonomy of curriculum learning approaches by hand, considering various classification criteria. We further build a hierarchical tree of curriculum learning methods using an agglomerative clustering algorithm, linking the discovered clusters with our taxonomy. At the end, we provide some interesting directions for future work.
\end{abstract}

{\section{Introduction}\label{sec:introduction}}

\noindent
{\bf Context and motivation.}
Deep neural networks have become the state-of-the-art approach in a wide variety of tasks, ranging from object recognition in images \citep{Krizhevsky-NIPS-2012,Simonyan-ICLR-14,Szegedy-CVPR-2015,He-CVPR-2016} and medical imaging \citep{Chen-MICCAI-2018,Ronneberger-MICCAI-2015,Kuo-PNAS-2019,burduja2020accurate} to text classification \citep{Brown-NeurIPS-2020,Devlin-NAACL-2019,Zhang-NIPS-2015,Kim-AAAI-2016} and speech recognition \citep{zhang2013denoising,Ravanelli-SLT-2018}. The main focus in this area of research is on building deeper and deeper neural architectures, this being the main driver for the recent performance improvements. For instance, the CNN model of Krizhevsky et al.~\citeyearpar{Krizhevsky-NIPS-2012} reached a top-5 error of $15.4\%$ on ImageNet \citep{Russakovsky-IJCV-2015} with an architecture formed of only 8 layers, while the more recent ResNet model \citep{He-CVPR-2016} reached a top-5 error of $3.6\%$ with 152 layers. While the CNN architecture has evolved over the last few years to accommodate more convolutional layers, reduce the size of the filters, and even eliminate the fully-connected layers, comparably less attention has been paid to improving the training process. 
An important limitation of the state-of-the-art neural models mentioned above is that examples are considered in a random order during training. Indeed, the training is usually performed with some variant of mini-batch stochastic gradient descent, the examples in each mini-batch being chosen randomly.

Since neural network architectures are inspired by the human brain, it seems reasonable to consider that the learning process should also be inspired by how humans learn. One essential difference from how machines are typically trained is that humans learn the basic (easy) concepts sooner and the advanced (hard) concepts later. This is basically reflected in all the curricula taught in schooling systems around the world, as humans learn much better when the examples are not randomly presented but are organized in a meaningful order. Using a similar strategy for training a machine learning model, we can achieve two important benefits: $(i)$ an increase of the convergence speed of the training process and $(ii)$ a better accuracy. A preliminary study in this direction has been conducted by Elman \citeyearpar{Elman-C-1993}. To our knowledge, Bengio et al.~\citeyearpar{Bengio-ICML-2009} are the first to formalize the easy-to-hard training strategies in the context of machine learning, proposing the \emph{curriculum learning} (CL) paradigm. This seminal work inspired many researchers to pursue curriculum learning strategies in various application domains, such as weakly supervised object localization \citep{Ionescu-CVPR-2016,shi2016weakly,tang2018attention}, object detection \citep{Chen_2015_ICCV,Li-BMVC-2017,sangineto2018self,Wang-ICPR-2018} and neural machine translation \citep{kocmi2017curriculum,zhang2018empirical,platanios2019competence,wang-etal-2019-dynamically} among many others. The empirical results presented in these works show the clear benefits of replacing the conventional training based on random mini-batch sampling with curriculum learning. Despite the consistent success of curriculum learning across several domains, this training strategy has not been adopted in mainstream works. This fact motivated us to write this survey on curriculum learning methods in order to increase the popularity of such methods. On another note, researchers proposed opposing strategies emphasizing harder examples, such as Hard Example Mining (HEM) \citep{shrivastava2016training,jesson2017cased,wang2018towards,zhou2020curriculum} or anti-curriculum \citep{pi2016self,braun2017curriculum}, showing improved results in certain conditions. 

\noindent
{\bf Contributions.}
Our first contribution is to formalize the existing curriculum learning methods under a single umbrella. This enables us to define a generic formulation of curriculum learning. Additionally, we link curriculum learning with the four main components of any machine learning approach: the data, the model, the task and the performance measure. We observe that curriculum learning can be applied on each of these components, all these forms of curriculum having a joint interpretation linked to loss function smoothing. Furthermore, we manually create a taxonomy of curriculum learning methods, considering orthogonal perspectives for grouping the methods: data type, task, curriculum strategy, ranking criterion and curriculum schedule. We corroborate the manually constructed taxonomy with an automatically built hierarchical tree of curriculum methods. In large part, the hierarchical tree confirms our taxonomy, although it also offers some new perspectives. While gathering works on curriculum learning and defining a taxonomy on curriculum learning methods, our survey is also aimed at showing the advantages of curriculum learning. Hence, our final contribution is to advocate the adoption of curriculum learning in mainstream works. 

\noindent
{\bf Related surveys.}
We are not the first to consider providing a comprehensive analysis of the methods employing curriculum learning in different applications. Recently, Narvekar et al.~\citeyearpar{narkevarJMLR20survey} survey the use of curriculum learning applied to reinforcement learning. They present a new framework and use it to survey and classify the existing methods in terms of their assumptions, capabilities and goals. They also investigate the open problems and suggest directions for curriculum RL research. While their survey is related to ours, it is clearly focused on RL research and, as such, is less general than ours. Directly relevant to our work is the recent survey of Wang et al.~\citeyearpar{WangPAMI20survey}. Their aim is similar to ours as they cover various aspects of curriculum learning including motivations, definitions, theories and several potential applications. We are looking at curriculum learning from a different view point and propose a generic formulation of curriculum learning. We also corroborate the automatically built hierarchical
tree of curriculum methods with the manually constructed taxonomy, allowing us to see curriculum learning from a new perspective. Furthermore, our review is more comprehensive, comprising nearly 200 scientific works. We strongly believe that having multiple surveys on the field will strengthen the focus and bring about the adoption of CL approaches in the mainstream research.

\noindent
{\bf Organization.} We provide a generic formulation of curriculum learning in Section~\ref{sec_curriculum}. We detail our taxonomy of curriculum learning methods in Section~\ref{sec_taxonomy}. We showcase applications of curriculum learning in Section~\ref{sec_applications} and we present the tree of curriculum approaches constructed by a hierarchical clustering approach in Section~\ref{sec_clustering}. Our closing remarks and directions of future study are provided in Section~\ref{sec_conclusion}.

\vspace{0.2cm}

\section{Curriculum Learning}
\label{sec_curriculum}

Mitchell \citeyear{Mitchell-MH-1997} proposed the following definition of machine learning:
\begin{definition}\label{def_ML}
A model $M$ is said to learn from experience $E$ with respect to some class of tasks $T$ and performance measure $P$, if its performance at tasks in $T$, as measured by $P$, improves with experience $E$.
\end{definition}

In the original formulation, Bengio et al.~\citeyearpar{Bengio-ICML-2009} proposed curriculum learning as a method to gradually increase the complexity of the data samples used during the training process. This is the most natural way to perform curriculum learning as it represents the most direct way of imitating how humans learn. Apparently, with respect to Definition~\ref{def_ML}, it may look that curriculum learning is about increasing the complexity of the experience $E$ during the training process. Most of the studies on curriculum learning follow this natural approach \citep{Bengio-ICML-2009,spitkovsky2009babysteps,Chen_2015_ICCV,zaremba2014learning,shi2015recurrent,pentina2015curriculum,Ionescu-CVPR-2016}. However, some studies propose to apply curriculum with respect to the other components in the definition of Mitchell \citeyear{Mitchell-MH-1997}. For instance, a series of methods proposed to gradually increase the modeling capacity of the model $M$ by adding neural units \citep{Karras-ICLR-2018}, by deblurring convolutional filters \citep{sinha2020curriculum} or by activating more units \citep{Morerio-ICCV-2017}, as the training process advances. Another set of methods relate to the class of tasks $T$, performing curriculum learning by increasing the complexity of the tasks \citep{zhang2017curriculum,lotter2017multi,sarafianos2017curriculum,florensa2017reverse,caubriere2019curriculum}. If we consider these alternative formulations from the perspective of the optimization problem, we can conclude that they are in fact equivalent. As pointed out by Bengio et al.~\citeyearpar{Bengio-ICML-2009}, the original formulation of curriculum learning can be viewed as a continuation method. The continuation method is a well-known approach in non-convex optimization \citep{allgower2003introduction}, which starts from a simple (smoother) objective function that is easy to optimize. Then, the objective function is gradually transformed into less smooth versions until it reaches the original (non-convex) objective function. In machine learning, we typically consider the objective function to be the performance measure $P$ in Definition~\ref{def_ML}. When we only use easy data samples at the beginning of the training process, we naturally expect that the model $M$ can reach higher performance levels faster. This is because the objective function should be smoother, as noted by Bengio et al.~\citeyearpar{Bengio-ICML-2009}. As we increase the difficulty of the data samples, the objective function should also become more complex. We highlight that the same phenomenon might apply when we perform curriculum over the model $M$ or the class of tasks $T$. For example, a model with lower capacity, e.g.,~a linear model, will inherently have a less complex, e.g.,~convex, objective function. Increasing the capacity of the model will also lead to a more complex objective. Linking curriculum learning to continuation methods allows us to see that applying curriculum with respect to the experience $E$, the model $M$, the class of tasks $T$ or the performance measure $P$ is leading to the same thing, namely to smoothing the loss function, in some way or another, in the preliminary training steps. While these forms of curriculum are somewhat equivalent, each bears its advantages and disadvantages. For example, performing curriculum with respect to the experience $E$ or the class of tasks $T$ may seem more natural. However, these forms of curriculum may require an external measure of difficulty, which might not always be available. Since Ionescu et al.~\citeyearpar{Ionescu-CVPR-2016} introduced a difficulty predictor for natural images, the lack of difficulty measures for this domain is no longer an issue. This fortunate situation is not often encountered in other domains. However, performing curriculum by gradually increasing the capacity of the model \citep{Karras-ICLR-2018,Morerio-ICCV-2017,sinha2020curriculum} does not suffer from this problem.

\begin{figure*}[ht]
\hspace{-0.01\linewidth}
\subfloat[General framework for data-level curriculum learning.]
{
	\includegraphics[width=0.526\linewidth]{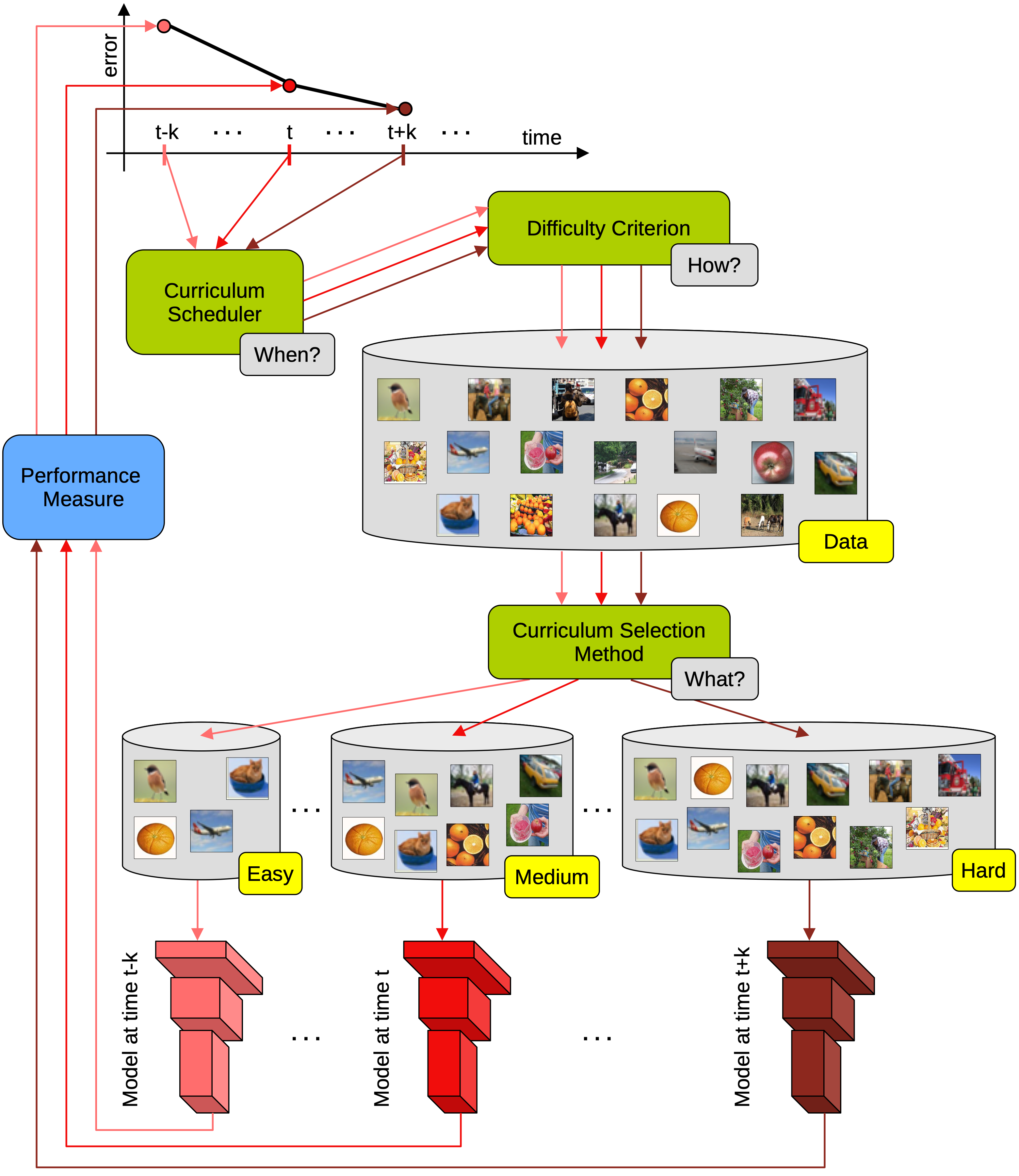}
    \label{fig_data_curr} 
}
\hspace{-0.02\linewidth}
\subfloat[General framework for model-level curriculum.]
{
 	\includegraphics[width=0.474\linewidth]{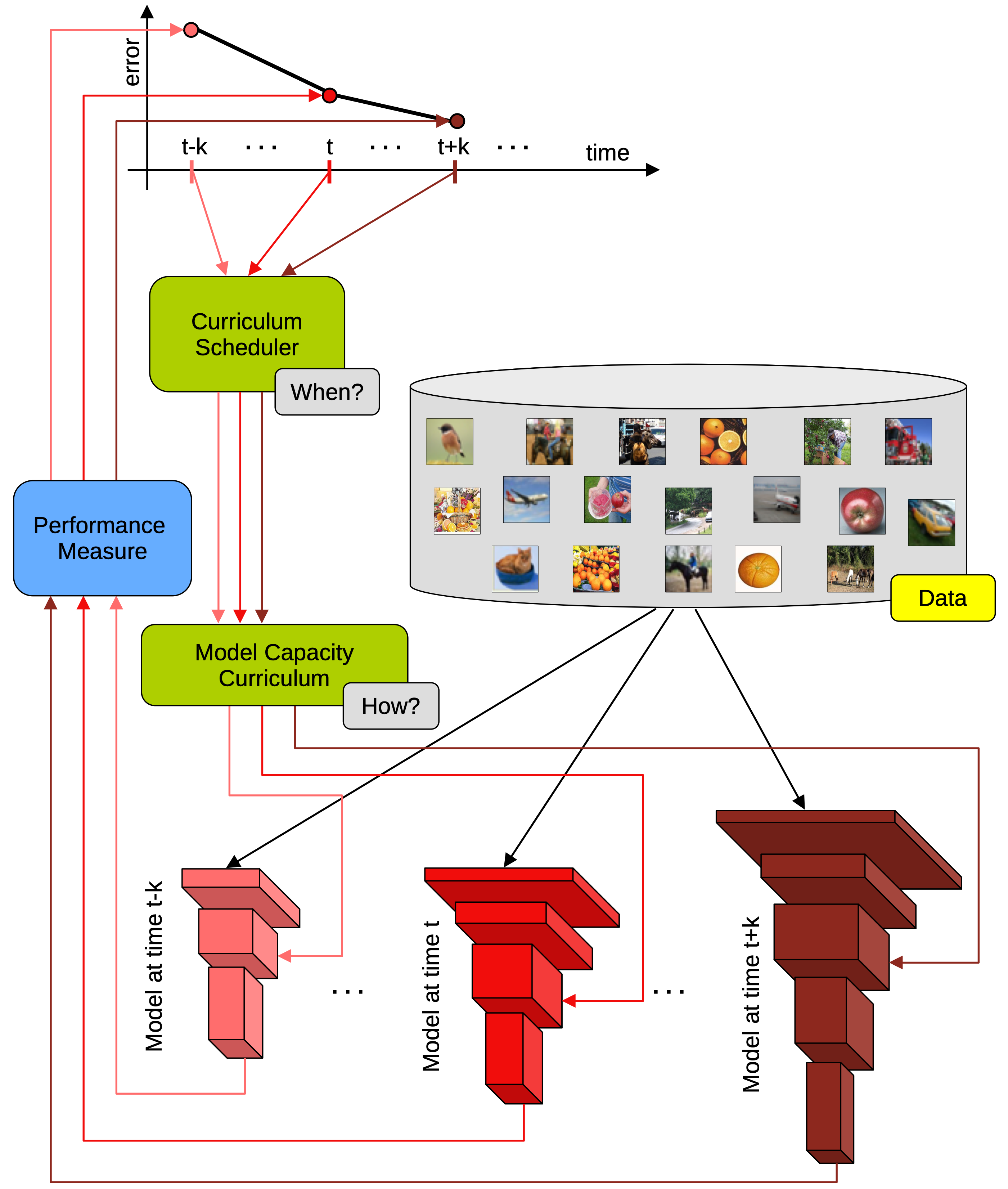}
    \label{fig_model_curr} 
}
\vspace{-0.1cm}
\caption{General frameworks for data-level and model-level curriculum learning, side by side. In both cases, $k$ is some positive integer. Best viewed in color.}
\label{fig_pipeline}
\end{figure*}

Figures~\ref{fig_data_curr} and~\ref{fig_model_curr} illustrate the general frameworks for curriculum learning applied at the data level and the model level, respectively. The two frameworks have two common elements: the curriculum scheduler and the performance measure. The scheduler is responsible for deciding when to update the curriculum in order to use the pace that gives the highest overall performance. Depending on the applied methodology, the scheduler may consider a linear pace or a logarithmic pace. Additionally, in self-paced learning, the scheduler can take into consideration the current performance level to find the right pace. When applying CL over data (see Figure~\ref{fig_data_curr}), a difficulty criterion is employed in order to rank the examples from easy to hard. Next, a selection method determines which examples should be used for training at the current time. Curriculum over tasks works in a very similar way. In Figure~\ref{fig_model_curr}, we observe that CL at the model level does not require a difficulty criterion. Instead, it requires the existence of a model capacity curriculum. This sets how to change the architecture or the parameters of the model to which all the training data is fed.

On another note, we remark that continuation methods can be seen as curriculum learning performed over the performance measure $P$ \citep{Pathak-ICMLA-2019}. However, this connection is not typically mentioned in literature. Moreover, continuation methods \citep{allgower2003introduction,Richter-TAC-1983,chow1991homotopy} were studied long before curriculum learning appeared \citep{Bengio-ICML-2009}. Research on continuation methods is therefore considered an independent field of study \citep{allgower2003introduction,chow1991homotopy}, not necessarily bound to its applications in machine learning \citep{Richter-TAC-1983}, as would be the case for curriculum learning.

\begin{algorithm}[t]
\caption{General curriculum learning algorithm}
\label{curriculum_alg}
\begin{algorithmic}[1]
 \renewcommand{\algorithmicrequire}{\textbf{}}
 \REQUIRE $M$ -- a machine learning model;
 \REQUIRE $E$ -- a training data set;
 \REQUIRE $P$ -- performance measure;
 \REQUIRE $n$ -- number of iterations / epochs;
 \REQUIRE $C$ -- curriculum criterion / difficulty measure;
 \REQUIRE $l$ -- curriculum level;
 \REQUIRE $S$ -- curriculum scheduler;
 
  \FOR{$t \in {1,2,...,n}$}
    \STATE $p \leftarrow P(M)$
	\IF {$S(t,p)$ = true}
	    \STATE $M, E, P \leftarrow C(l, M, E, P)$
	\ENDIF
	\STATE $E^* \leftarrow select(E)$
	\STATE $M \leftarrow train(M,E^*,P)$
   \ENDFOR
\end{algorithmic}
\end{algorithm}

\begin{sloppypar}
We propose a generic formulation of curriculum learning that should encompass the equivalent forms of curriculum presented above. 
Algorithm \ref{curriculum_alg} illustrates the common steps involved in the curriculum training of a model $M$ on a data set $E$. It requires the existence of a curriculum criterion $C$, i.e.,~a methodology of how to determine the ordering, and a level $l$ at which to apply the curriculum, e.g.,~data level, model level, or performance measure level. The traditional curriculum learning approach enforces an easy-to-hard re-ranking of the examples, with the criterion, or the difficulty metric, being task-dependent, such as the use of shape complexity for images \citep{Bengio-ICML-2009,DBLP:conf/eccv/DuanZ0YNG20}, grammar properties for text \citep{kocmi2017curriculum,liu2018curriculum} and signal-to-noise ratio for audio \citep{braun2017curriculum,ranjan2017curriculum}. Nevertheless, more general methods can be applied when generating the curriculum, e.g.,~supervising the learning by a teacher network (teacher-student) \citep{jiang2018mentornet,wu2018learning,kim2018screenernet} or taking into consideration the learning progress of the model (self-paced learning) \citep{Kumar-NIPS-2010,jiang2014self,zhao2015self,Jiang-AAAI-2015,zhang2015self}. The easy-to-hard ordering can also be applied when multiple tasks are involved, determining the best order to learn the tasks to maximize the final result \citep{pentina2015curriculum,zhang2017curriculum,lotter2017multi,sarafianos2017curriculum,florensa2017reverse,matiisen2019teacher}. A special type of methodology is when the curriculum is applied at the model level, adapting various elements of the model during its training \citep{Morerio-ICCV-2017,Karras-ICLR-2018,wu2018learning,sinha2020curriculum}. Another key element of any curriculum strategy is the scheduling function $S$, which specifies when to update the training process. The curriculum learning algorithm is applied on top of the traditional learning loop used for training machine learning models. At step $11$, we compute the current performance level $p$, which might be used by the scheduler $S$ to determine the right moment for applying the curriculum. We note that the scheduler $S$ can also determine the pace solely based on the current training iteration/epoch $t$. Steps $11$-$13$ represent the part specific to curriculum learning. At step $13$, the curriculum criterion alternatively modifies the data set $E$ (e.g., by sorting it in increasing order of difficulty), the model $M$ (e.g., by increasing its modeling capacity), or the performance measure $P$ (e.g., by unsmoothing the objective function). We hereby emphasize once again that the criterion function $C$ operates on $M$, $E$, or $P$, according to the value of $l$. At the same time, we should not exclude the possibility to employ curriculum at multiple levels, jointly. At step $14$, the training loop performs a standard operation, i.e., selecting a subset $E^* \subseteq E$, e.g.,~a mini-batch, which is subsequently used at step $15$ to update the model $M$. It is important to note that, when the level $l$ is the data level, the data set $E$ is organized at step $13$ such that the selection performed at step $14$ chooses a subset with the proper level of difficulty for the current time $t$. In the context of data-level curriculum, common approaches for selecting the subset $E^*$ are batching \citep{Bengio-ICML-2009,lee2011learning,Choi2019PseudoLabelingCF}, weighting \citep{Kumar-NIPS-2010,zhang2015self,liang2016learning} and sampling \citep{jiang2014easy,Li-BMVC-2017,jesson2017cased}. Yet, other specific iterative \citep{spitkovsky2009babysteps,pentina2015curriculum,Gong-TIP-2016} or continuous methodologies have been proposed \citep{shi2015recurrent,Morerio-ICCV-2017,bassich2019continuous} in the literature.
\end{sloppypar}

\section{Taxonomy of Curriculum Learning Methods}
\label{sec_taxonomy}

We next present a multi-perspective categorization of the papers reviewed in this survey. Although curriculum learning approaches can be divided into different types with respect to the components involved in Definition~\ref{def_ML}, this categorization is extremely unbalanced, as most of the proposed methods are actually based on data-level curriculum (see Table~\ref{tab_taxonomy}). To this end, we devise a more balanced partitioning formed of seven categories, which stem from the different assumptions, model requirements and training methodologies applied in each work. The seven categories representing various forms of curriculum learning (CL) are: vanilla CL, self-paced learning (SPL), balanced CL (BCL), self-paced CL (SPCL), teacher-student CL, implicit CL (ICL) and progressive CL (PCL). Reasonably, the proposed taxonomy must not be considered as a sharp and exhaustive categorization of different curriculum solutions. On the contrary, hybrid and smooth implementations are also quite common, as can be noticed in Table~\ref{tab_taxonomy}. Besides classifying the reviewed papers according the aforementioned categories, we consider alternative categorization criteria, such as the application domain or the addressed task. Together, these criteria determine the multi-perspective categorization of the reviewed articles, which is presented in Table~\ref{tab_taxonomy}.


\noindent\textbf{Vanilla CL} was introduced in 2009 by \citeauthor{Bengio-ICML-2009}, who proved that machine learning models are improving their performance levels when fed with increasingly difficult samples during training. Vanilla CL, or simply CL in the rest of this paper, is where curriculum is used as the only rule-based criterion for sample selection. 
In general, CL exploits a set of a priori rules to discriminate between easy and hard examples. The seminal paper of Bengio et al.~\citeyearpar{Bengio-ICML-2009} is a clear example where geometric shapes are fed from basic to complex to the model during training. Another clear example is proposed in \citep{spitkovsky2009babysteps}, where the authors exploit the length of sequences, claiming that longer sequences are harder to predict than shorter ones.

\noindent\textbf{Self-paced learning} (SPL) differs from the previous category in the way samples are being evaluated. More specifically, the main difference with respect to Vanilla CL is related to the order in which the samples are fed to the model. In SPL, such order is not known a priori, but computed with the respect to the model's own performance, and therefore, it may vary during training. Indeed, the difficulty is measured repeatedly during training, altering the order of samples in the process. In \citep{Kumar-NIPS-2010}, for instance, the likelihood of the prediction is used to rank the samples, while in \citep{lee2011learning}, the \textit{objectness} is considered to define the training order.

\noindent\textbf{Balanced curriculum (BCL)} is based on the intuition that, in addition to the traditional CL training criteria, samples have to be diverse enough while being proposed to the model. This category introduces multiple ordering criteria. According to the difficulty criterion, the model is fed with easy samples first, and then, as the training progresses, harder samples are added. At the same time, the selection of samples has to be balanced under additional constraints, such as constraints that ensure diversity across image regions \citep{zhang2015self} or classes \citep{soviany2020curriculum}.


\begin{sloppypar}
\noindent\textbf{Self-paced curriculum learning (SPCL)}. 
In the introductory part of this section, we clearly stated that a possible overlap between categories is not only possible, but actually frequent. SPL and CL, however, in our opinion require a specific mention, since many works are drawing jointly from the two categories. To this end, we can specifically identify SPCL, a paradigm where predefined criteria and learning-based metrics are jointly used to define the training order of samples. 
This paradigm has been first presented by Jiang et al.~\citeyearpar{Jiang-AAAI-2015} and applied to matrix factorization and multimedia event detection. It has also been exploited in other tasks such as weakly-supervised object segmentation in videos \citep{zhang2017spftn} or person re-identification \citep{ma2017self}.
\end{sloppypar}

\noindent\textbf{Progressive CL (PCL)} refers to the task in which the curriculum is not related to the difficulty of every single sample, but is configured instead as a progressive mutation of the model capacity or task settings. In principle, PCL does not implement CL with respect to the sample order (the samples are indeed provided to the model in a traditional random order), instead applying the curriculum concept to a connected task or to a specific part of the network, resulting in an easier task at the beginning of the training, which gets harder towards the end. An example is the Curriculum Dropout of Morerio et al.~\citeyearpar{Morerio-ICCV-2017}, where a monotonic function is devised to decrease the probability of the dropout during training. The authors claim that, at the beginning of the training, dropout will be weak and should progressively increase to significantly improve performance levels. Another example for this category is the approach proposed in \citep{Karras-ICLR-2018}, which progressively grows the capacity of Generative Adversarial Networks to obtain high-quality results.

\noindent\textbf{Teacher-student CL} splits the training into two tasks, a model that learns the principal task (student) and an auxiliary model (teacher) that determines the optimal learning parameters for the student. In this specific architecture, the curriculum is implemented via a network that applies the policy on a student model that will eventually provide the final inference. Such an approach has been first proposed by Kim and Choi \citeyearpar{kim2018screenernet} in a deep reinforcement learning fashion, and then, reformulated in later works \citep{jiang2018mentornet,Hacohen2019OnTP,zhang2019automatic}.

\noindent\textbf{Implicit CL} is when CL has been applied without specifically building a curriculum, like when organizing the data accordingly. Instead, the easy-to-hard schedule can be regarded as a side effect of a specific training methodology. For example, Sinha et al.~\citeyearpar{sinha2020curriculum} propose to gradually deblur convolutional activation maps during training. This procedure can be seen as a sort of curriculum mechanism where, at first, the network exhibits a reduced learning capacity and gets more complex with the prosecution of the training. Another example is proposed by Almeida et al.~\citeyearpar{almeida2020low}, where the goal is to reduce the number of labeled samples to reach a certain classification performance. To this end, unsupervised training is performed on the raw data to determine a ranking based on the informativeness of each example.

\noindent\textbf{Category overlap.} As mentioned earlier, we do not view the proposed categories as disjoint, but rather as pools of approaches that often intersect with each other. Perhaps the most relevant example in this direction is the combination of CL and SPL which was already adopted in multiple works from the literature and led to the development of SPCL. However, this example is not singular. As shown in Table~\ref{tab_taxonomy}, a few reported works are leveraging aspects from multiple categories. For instance, BCL has been employed together with multiple SPL \citep{jiang2014self, sachan2016easy, ren2017robust} and teacher-student approaches \citep{zhao2021automatic,zhao2020egdcl}. A recent example in this direction is the work of Zhang et al.~\citeyearpar{zhang2021flexmatch}, where a self-paced technique is proposed to improve image classification. This method is however sided by a threshold-based system that mitigates the attitude of an SPL method to sample in an unbalanced manner, for this reason being classified as SPL and BCL. Another interesting combination is the use of teacher-student frameworks together with complexity based approaches \citep{kim2018screenernet,Hacohen2019OnTP,huang2019self}. For example, Kim and Choi \citeyearpar{kim2018screenernet} train the teacher and student networks together, using an SPL approach based on the loss of the student.

\noindent\textbf{Related methodologies.} Besides the standard easy-to-hard approach employed in CL, other strategies differ in the way they build the curriculum. In this direction, Shrivastava et al.~\citeyearpar{shrivastava2016training} employ a Hard Example Mining (HEM) strategy for object detection which emphasizes difficult examples, i.e.,~examples with higher loss. Braun et al.~\citeyearpar{braun2017curriculum} utilize anti-curriculum learning (Anti-CL) for automatic speech recognition systems under noisy environments, using the signal-to-noise ratio to create a hard-to-easy ordering. On another note, active learning (AL) setups do not focus on the difficulty of the examples, but on the uncertainty. Chang et al.~\citeyearpar{chang2017active} claim that SPL and HEM might work well in different scenarios, but sorting the examples based on the level of uncertainty, in an AL fashion, might provide a general solution for achieving higher quality results. Tang and Huang \citeyearpar{tang2019self} combine active learning and SPL, creating an algorithm that jointly considers the potential value of the examples for improving the overall model and the difficulty of the instances.

\noindent\textbf{Other elements of curriculum learning.} Besides the methodology-based categorization, curriculum techniques employ different criteria for building the curriculum, multiple scheduling approaches, and can be applied on different levels (data, task, or model).

Traditional easy-to-hard CL techniques build the curriculum by taking into consideration the difficulty of the examples or the tasks. This can be manually labeled using human annotators \citep{pentina2015curriculum,Ionescu-CVPR-2016,lotfian2019curriculum,jimenez2019medical,wei2020learn} or automatically determined using predefined task or domain-dependent difficulty measures. For example, the length of the sentence \citep{spitkovsky2009babysteps,cirik2016visualizing,kocmi2017curriculum,subramanian2017adversarial,zhang2018empirical} or the term frequency \citep{Bengio-ICML-2009,kocmi2017curriculum,liu2018curriculum} are used in NLP, while the size of the objects is employed in computer vision \citep{shi2016weakly,soviany2019curriculum}. Another solution for automatically generating difficulty scores is to consider the results of a different network on the training examples \citep{Gong-TIP-2016,zhang2018empirical,Hacohen2019OnTP} or to use a difficulty estimation model \citep{Ionescu-CVPR-2016,wang2018towards,Soviany-WACV-2020}. Compared to the standard predefined ordering, teacher-student models usually generate the curriculum dynamically, taking into consideration the progress of the student network under the supervision of the teacher \citep{jiang2018mentornet,wu2018learning,kim2018screenernet,zhang2019automatic}. The learning progress is also used in SPL, where the easy-to-hard ordering is enforced using the current value of the loss function \citep{Kumar-NIPS-2010,jiang2014self,zhao2015self,Jiang-AAAI-2015,zhang2015self,fan2016self,pi2016self,li2016multi,zhou2018deep,gong2018decomposition,ma2018convergence,sun2020fspmtl}. Similarly, in reinforcement learning setups, the order of tasks is determined so as to maximize a reward function \citep{qu2018curriculum,pmlr-v100-klink20a,narvekar2016source}.

Multiple scheduling approaches are employed when building a curriculum strategy. Batching refers to the idea of splitting the training set into subsets and commencing learning from the easiest batch \citep{Bengio-ICML-2009,lee2011learning,Ionescu-CVPR-2016,Choi2019PseudoLabelingCF}. As the training progresses, subsets are gradually added, enhancing the training set. The ``easy-then-hard'' strategy is similar, being based on splitting the original set into subgroups. Still, the training set is not augmented, but each group is used distinctively for learning \citep{Bengio-ICML-2009,Chen_2015_ICCV,sarafianos2017curriculum}. In the sampling technique, training examples are selected according to certain difficulty constraints \citep{jiang2014easy,Li-BMVC-2017,jesson2017cased}. Weighting appends the difficulty to the learning procedure, biasing the models towards certain examples, considering the training stage \citep{Kumar-NIPS-2010,zhang2015self,liang2016learning}. Another scheduling strategy for selecting the easier samples for learning is to remove hard examples \citep{wang-etal-2019-dynamically,castells2020superloss,wang2020learning}. Curriculum methods can also be scheduled in different stages, with each stage focusing on a distinct task \citep{narvekar2016source,zhang2017curriculum,lotter2017multi}. Aside from these categories, we also define the continuous \citep{shi2015recurrent,Morerio-ICCV-2017,bassich2019continuous} and iterative \citep{spitkovsky2009babysteps,pentina2015curriculum,Gong-TIP-2016,sachan2016easy} scheduling, for specific methods which adapt more general approaches.

\section{Applications of Curriculum Learning}
\label{sec_applications}

In this section, we perform an extensive exploration of the curriculum learning literature, briefly describing each paper. The works are grouped at two levels, first by domain, then by task, with similar approaches being presented one after another in order to keep the logical flow of the reading. By choosing this ordering, we enable the readers to find the papers which address their field of interest, while also highlighting the development of curriculum methodologies in each domain or task.
Table~\ref{tab_taxonomy} illustrates each distinctive element of the  curriculum learning strategies for the selected papers.

\begin{table*}[!th]
\setlength\tabcolsep{2pt}
\caption{Multi-perspective taxonomy of curriculum learning methods.} 
\label{tab_taxonomy}
\centering
\begin{tabular}{|p{4cm}|p{1.5cm}|p{3cm}|p{1.6cm}|p{2.3cm}|p{2cm}|p{1.5cm}|}
\hline
\textbf{Paper}                           & \textbf{Domain} & \textbf{Tasks}                                                                              & \textbf{Method}     & \textbf{Criterion}                                                                                              & \textbf{Schedule}                 & \textbf{Level} \\ \hline
Bengio et al. \citeyearpar{Bengio-ICML-2009}                  & CV, NLP         & shape recognition, next best word                                                           & CL                    & shape complexity, word frequency                                                                                & easy-then-hard, batching             & data           \\ \hline
Spitkovsky et al. \citeyearpar{spitkovsky2009babysteps}    & NLP             & grammar induction                                                              & CL                    & sentence length                                                                                                 & iterative                            & data           \\ \hline
Kumar et al. \citeyearpar{Kumar-NIPS-2010}                   & CV, NLP         & noun phrase coreference, motif finding, handwritten digits recognition, object localization  & SPL                   & loss                                                                                                            & weighting                            & data           \\ \hline
Kumar et al. \citeyearpar{kumar2011learning}                   & CV              & semantic segmentation                                                                       & SPL                   & loss                                                                                                            & weighting                            & data           \\
\hline
\end{tabular}
\end{table*}

\begin{table*}[!th]
\setlength\tabcolsep{2pt}
\centering
\begin{tabular}{|p{4cm}|p{1.5cm}|p{3cm}|p{1.6cm}|p{2.3cm}|p{2cm}|p{1.5cm}|}
\hline
Amodei et al. \citeyearpar{amodei2016deep}                  & Speech          & speech recognition                                                                          & CL                    & length of utterance                                                                                             & iterative                            & data           \\ \hline
Liang et al. \citeyearpar{liang2016learning}                 & CV              & detection                                                                                   & SPCL                  & term frequency in video metadata, loss                                                                          & weighting                            & data           \\ \hline

Narvekar et al. \citeyearpar{narvekar2016source}               & ML              & RL, games                                                                                   & CL                    & learning progress                                                                                               & stages                               & data, task     \\ \hline
Graves et al. \citeyearpar{graves2016hybrid}                & ML              & graph traversal task                                                                        & CL                    & number of nodes, edges, steps & iterative                            & data           \\ \hline
Morerio et al. \citeyearpar{Morerio-ICCV-2017}                 & CV              & classification                                                                              & PCL                   & dropout                                                                                                         & continuous                           & model          \\
\hline

\end{tabular}
\end{table*}

\begin{table*}[!th]
\setlength\tabcolsep{2pt}
\centering
\begin{tabular}{|p{4cm}|p{1.5cm}|p{3cm}|p{1.6cm}|p{2.3cm}|p{2cm}|p{1.5cm}|}
\hline
Wang and Vasconcelos \citeyearpar{wang2018towards}           & CV              & classification, scene recognition                                                           & CL, HEM               & difficulty estimator network                                                                                    & sampling                             & data           \\ \hline
Kim and Choi \citeyearpar{kim2018screenernet}                        & CV              & classification                                                                              & SPL, teacher-student  & loss                                                                                                            & weighting                            & data           \\ \hline
Weinshall et al. \citeyearpar{Weinshall2018CurriculumLB}               & CV              & classification                                                                              & CL                    & transfer learning                                                                                               & stages                               & data           \\ \hline
Zhou and Bilmes \citeyearpar{zhou2018minimax}                & CV              & classification                                                                              & BCL, HEM             & loss                                                                                                            & iterative                            & data           \\
\hline
\end{tabular}
\end{table*}

\begin{table*}[!th]
\setlength\tabcolsep{2pt}
\centering
\begin{tabular}{|p{4cm}|p{1.5cm}|p{3cm}|p{1.6cm}|p{2.3cm}|p{2cm}|p{1.5cm}|}
\hline
Tay et al. \citeyearpar{tay-etal-2019-simple}                      & NLP             & reading comprehension                                                                       & BCL                  & answerability, understandability                                                                                & batching                             & data           \\ \hline
Lotfian and Busso \citeyearpar{lotfian2019curriculum}             & Speech          & speech emotion recognition                                                                  & CL                    & human-assessed ambiguity                                                                                         & batching                             & data           \\ \hline
Zheng et al. \citeyearpar{zheng2019autoencoder}          & Speech          & DA, speaker verification                                                                    & CL                    & domain                                                                                                          & batching, iterative                  & data           \\ \hline
Caubriere et al. \citeyearpar{caubriere2019curriculum}        & Speech          & language understanding                                                                      & CL                    & subtask specificity                                                                                             & stages                               & task           \\ \hline
Zhang et al. \citeyearpar{zhang2019automatic}              & Speech          & digital modulation classification                                                           & teacher-student       & loss                                                                                                            & weighting                            & data           \\
\hline

\end{tabular}
\end{table*}

\begin{table*}[!th]
\setlength\tabcolsep{2pt}
\centering
\begin{tabular}{|p{4cm}|p{1.5cm}|p{3cm}|p{1.6cm}|p{2.3cm}|p{2cm}|p{1.5cm}|}
\hline
Duan et al. \citeyearpar{DBLP:conf/eccv/DuanZ0YNG20}                    & CV              & shape reconstructions                                                                       & CL                    & surface accuracy, sample complexity                                                                             & weighting                            & data           \\ \hline
Zhou et al. \citeyearpar{zhou2020uncertainty}                     & NLP             & machine translation                                                                         & CL                    & uncertainty                                                                                                     & batching                             & data           \\ \hline
Guo et al. \citeyearpar{guo2020fine}                     & NLP             & machine translation                                                                         & CL                    & task difficulty                                                                                                 & stages                               & task           \\ \hline
Liu et al. \citeyearpar{liu2020task}                    & NLP             & machine translation                                                                         & CL                    & task difficulty                                                                                                 & iterative                            & task           \\ \hline
Wang et al. \citeyearpar{wang2020learning}                     & NLP             & machine translation                                                                         & CL                    & performance                                                                                                     & discard difficult examples, sampling & data           \\
\hline

\end{tabular}
\end{table*}

\begin{table*}[!th]
\setlength\tabcolsep{2pt}
\centering
\begin{tabular}{|p{4cm}|p{1.5cm}|p{3cm}|p{1.6cm}|p{2.3cm}|p{2cm}|p{1.5cm}|}
\hline
 
Zhan et al. \citeyearpar{zhan2021meta}                & 
NLP              & machine translation, DA                                                                      & CL                  & domain specificity                                                                            & sampling                           & task \\ \hline

Chang et al. \citeyearpar{chang2021does}              & 
NLP              & data-to-text generation                                                                      & CL                  & sentence length, word rarity, Damerau-Levenshtein distance, soft edit distance                                                                            & sampling                           & data \\ \hline

Zhang et al. \citeyearpar{zhang2021enhancing}                & 
Audio-visual              & action recognition, sound recognition                                                                      & teacher-student                  & task difficulty                                                                            & stages                           & task \\ \hline

Jafarpour et al. \citeyearpar{jafarpour2021active}              & 
NLP              & named entity recognition                                                                      & CL, AL                  & difficulty -- linguistic features, informativeness                                                                            & iterative                           & data \\ \hline

Burduja and Ionescu \citeyearpar{Burduja-ICIP-2021}              & 
Medical              & image registration                                                                      & CL, PCL                  & image sharpness, dropout, Gaussian kernel                                                                            & continuous                           & data, model \\
\hline

\end{tabular}
\end{table*}

\subsection{Multi-domain approaches}

In the first part of this section, we focus on the general curriculum learning solutions that have been tested in multiple domains. Two of the main works in this category are the papers that first formulated the vanilla curriculum learning and the self-paced learning paradigms. These works highly influenced the progress of easy-to-hard learning strategies and led to multiple approaches, which have been successfully employed in all domains and in a wide range of tasks.

Bengio et al.~\citeyearpar{Bengio-ICML-2009} introduce a set of easy-to-hard learning strategies for automatic models, referred to as Curriculum Learning (CL). The idea of presenting the examples in a meaningful order, starting from the easiest samples, then gradually introducing more complex ones, was inspired by the way humans learn. To show that automatic models benefit from such a training strategy, achieving faster convergence, while finding a better local minimum, the authors conduct multiple experiments. They start with toy experiments with a convex criterion in order to analyze the impact of difficulty on the final result. They find that, in some cases, easier examples can be more informative than more complex ones. Additionally, they discover that feeding a perceptron with the samples in increasing order of difficulty performs better than the standard random sampling approach or than a hard-to-easy methodology (anti-curriculum). Next, they focus on shape recognition, generating two artificial data sets: BasicShapes and GeomShapes, with the first one being designed to be easier, with less variability in terms of shape. They train a neural network on the easier set until a switch epoch when they start training on the GeomShapes set. The evaluation is conducted only on the difficult data, with the curriculum approach generating better results than the standard training method. The methodology above can be considered an adaptation of transfer learning, where the network was pre-trained on a similar, but easier, data set. Finally, the authors conduct language modeling experiments for predicting the best word which could follow a sequence of words in correct English. The curriculum strategy is built by iterating over Wikipedia and selecting the most frequent 5000 words from the vocabulary at each step. This vocabulary enhancement method compares favorably to conventional training. Still, their experiments are constructed in a way that enables the easy and the difficult examples to be easily separated. In practice, finding a way to rank the training examples can be a complex task.   

\begin{sloppypar}
Starting from the intuition of Bengio et al.~\citeyearpar{Bengio-ICML-2009}, Kumar et al.~\citeyearpar{Kumar-NIPS-2010} update the vanilla curriculum learning methodology and introduce self-paced learning (SPL), another training strategy that suggests presenting the training examples from easy to hard. The main difference from the standard curriculum approach is the method of computing the difficulty. CL assumes the existence of some external, predefined intuition, which can guide the model through the learning process. Instead, SPL takes into consideration the learning progress of the model in order to choose the next best samples to be presented. The method of Kumar et al.~\citeyearpar{Kumar-NIPS-2010} is an iterative approach which, at each step, jointly selects the easier samples and updates the parameters. The easiness is regarded as how facile is to predict the correct output, i.e., which examples have a higher likelihood to determine the correct output. The easy-to-hard transition is determined by a weight that is gradually increased to introduce more (difficult) examples in the later iterations, eventually considering all samples.
\end{sloppypar}

Li et al.~\citeyearpar{li2016multi} claim that standard SPL approaches are limited by the high sensitivity to initialization and the difficulty of finding the moment to terminate the incremental learning process. To alleviate these problems, the authors propose decomposing the objective into two terms, the loss and the self-paced regularizer, tackling the problem as the compromise between these two objectives. By reformulating the SPL as a multi-objective task, a multi-objective evolutionary algorithm can be employed to jointly optimize the two objectives and determine the right pace parameter.

Fan et al.~\citeyearpar{fan2016self} introduce the self-paced implicit regularizer, a group of new regularizers for SPL that is deduced from a robust loss function. SPL highly depends on the objective functions in order to obtain better weighting strategies, with other methods usually relying on artificial designs for the explicit form of SPL regularizers. To prove the correctness and effectiveness of implicit regularizers, the authors implement their framework on both supervised and unsupervised tasks, conducting matrix factorization, clustering and classification experiments.

Li et al.~\citeyearpar{li2016self} apply a self-paced methodology on top of a multi-task learning framework. Their algorithm takes into consideration both the complexity of the task and the difficulty of the examples in order to build the easy-to-hard schedule. They introduce a task-oriented regularizer to jointly prioritize tasks and instances. It contains the negative $l_1$-norm that favors the easy instances over the hard ones per task, together with an adaptive $l_{2,1}$-norm of a matrix, which favors easier tasks over the hard ones.

Li et al.~\citeyearpar{li2017self-b} present an SPL approach for learning a multi-label model. During training, they compute and use the difficulties of both instances and labels, in order to create the easy-to-hard ordering. Furthermore, the authors provide a general method for finding the appropriate self-paced functions. They experiment with multiple functions for the self-paced learning schemes, e.g.,~sigmoid, atan, exponential and tanh. Experiments on two image data sets and one music data set show the superiority of the SPL methodology over conventional training.

A thorough analysis of the SPL methodology is performed by Gong et al.~\citeyearpar{gong2018decomposition} in order to determine the right moment to optimally stop the incremental learning process. They propose a multi-objective self-paced method that jointly optimizes the loss and the regularizer. To optimize the two objectives, the authors employ a decomposition-based multi-objective particle swarm algorithm together with a polynomial soft weighting regularizer.

Jiang et al.~\citeyearpar{Jiang-AAAI-2015} consider that the standard curriculum learning and self-paced learning algorithms do not capture the full potential of the easy-to-hard strategies. On the one hand, curriculum learning uses a predefined curriculum and does not take into consideration the training progress. On the other hand, self-paced learning only relies on the learning progress, without using any prior knowledge. To overcome these problems, the authors introduce self-paced curriculum learning (SPCL), a learning methodology that combines the merits of CL and SPL, using both prior knowledge and the training progress. The method takes a predefined curriculum as input, guiding the model to the examples that should be visited first. The learner takes into consideration this knowledge while updating the curriculum to the learning objective, in an SPL manner. The SPCL approach was tested on two tasks: matrix factorization and multimedia event detection.

\begin{sloppypar}
Ma et al.~\citeyearpar{ma2017self} borrow the instructor-student-collaborative intuition from SPCL and introduce a self-paced co-training strategy. They extend the traditional SPL approach to the two-view scenario, by adding importance weights for the views on top of the corresponding regularizer. The algorithm uses a ``draw with replacement'' methodology, i.e., previously selected examples from the pool are kept only if the value of the loss is lower than a fixed threshold. To test their approach, the authors conduct extensive text classification and person re-identification experiments.
\end{sloppypar}

Wu et al.~\citeyearpar{wu2018learning} propose another easy-to-hard strategy for training automatic models: the teacher-student framework. On the one hand, teachers set goals and evaluate students based on their growth, assigning more difficult tasks to the more advanced learners. On the other hand, teachers improve themselves, acquiring new teaching methods and better adjusting the curriculum to the students' needs. For this, the authors propose a model in which the teacher network learns to generate appropriate learning objectives (loss functions), according to the progress of the student. Furthermore, the teacher network self-improves, its parameters being optimized during the teaching process. The gradient-based optimization is enhanced by smoothing the task-specific quality measure of the student and by reversing the stochastic gradient descent training process of the student model. 

\subsection{Computer vision}

All types of easy-to-hard learning strategies have been successfully employed in a wide range of computer vision problems. For the standard curriculum approach, various difficulty metrics for computing the complexity of the training examples have been proposed. Chen and Gupta~\citeyearpar{Chen_2015_ICCV} consider the source of the image to be related to the complexity, Soviany et al.~\citeyearpar{Soviany-CEFRL-2018} use object-related statistics such as the number or the average size of the objects, and Ionescu et al.~\citeyearpar{Ionescu-CVPR-2016} build an image complexity estimator. Furthermore, model \citep{sinha2020curriculum} and task-based approaches \citep{pentina2015curriculum} have also been explored to solve vision problems.

\vspace{0.2cm}
\noindent
\textbf{Multiple tasks.}
Chen and Gupta~\citeyearpar{Chen_2015_ICCV} introduce one of the first curriculum frameworks for computer vision. They use web images to train a convolutional neural network in a curriculum fashion. They collect information from Google and Flickr, arguing that Flickr images are noisier, thus more difficult. Starting from this observation, they build the curriculum training in two-stages: first they train the model on the easy Google images, then they fine-tune it on the more difficult Flickr examples. Furthermore, to smooth the difficulty of very hard samples, the authors impose constraints during the fine-tuning step, based on similarity relationships across different categories.

Ionescu et al.~\citeyearpar{Ionescu-CVPR-2016} measure the complexity of an image as the human response time required for a visual search task. Using human annotations, they build a regression model which can automatically estimate the difficulty score of a certain image. Based on this measure, they conduct curriculum learning experiments on two different tasks: weakly-supervised object localization and semi-supervised object classification, showing both the superiority of the easy-to-hard strategy and the efficiency of the estimator.

\begin{sloppypar}
Compared to the approach of Chen and Gupta~\citeyearpar{Chen_2015_ICCV}, the prior knowledge generated by the difficulty estimator of Ionescu et al.~\citeyearpar{Ionescu-CVPR-2016} is computed, thus being more general. Soviany~\citeyearpar{soviany2020curriculum} uses this estimator to build a curriculum sampling approach that addresses the problem of imbalance in fully annotated data sets. The author augments the easy-to-hard sampling strategy from \citep{Soviany-WACV-2020} with a new term that captures the diversity of the examples. The total number of objects in each class from the previously selected examples is counted in order to emphasize less-visited classes.
\end{sloppypar}

Wang and Vasconcelos~\citeyearpar{wang2018towards} introduce realistic predictors, a new class of predictors that estimate the difficulty of examples and reject samples considered too hard. They build a framework for the classification task in which the difficulty is computed using a network (HP-Net) that is jointly trained with the classifier. The two networks share the same inputs and are trained in an adversarial fashion, i.e.,~while the classifier improves its predictions, the HP-Net perfects its hardness scores. The softmax probabilities of the classifier are used to tune the HP-Net, using a variant of the standard cross-entropy as the loss. The difficulty score is then used to build a new training set by removing the most difficult examples from the data set.

Saxena et al.~\citeyearpar{saxena2019data} employ a different approach, using data parameters to automatically generate the curriculum to be followed by the model. They introduce these learnable parameters both at the sample level and at the class level, measuring their importance in the learning process. Data parameters are automatically updated at each iteration together with the model parameters, by gradient descent, using the corresponding loss values. Experiments show that, in noisy conditions, the generated curriculum follows indeed the easy-to-hard strategy, prioritizing clean examples at the beginning of the training.

Sinha et al.~\citeyearpar{sinha2020curriculum} introduce a curriculum by smoothing approach for convolutional neural networks, by convolving the output activation maps of each convolutional layer with a Gaussian kernel. During training, the variance of the Gaussian kernel is gradually decreased, thus allowing increasingly more high-frequency data to flow through the network. As the authors claim, the first stages of the training are essential for the overall performance of the network, limiting the effect of the noise from untrained parameters by setting a high standard deviation for the kernel at the beginning of the learning process.

Castells et al.~\citeyearpar{castells2020superloss} introduce a super loss approach to self-supervise the training of automatic models, similar to SPL. The main idea is to append a novel loss function on top of the existing task-dependent loss to automatically lower the contribution of hard samples with large losses. The authors claim that the main contribution of their approach is that it is task-independent, and prove the efficiency of their method using extensive experiments.

\vspace{0.2cm}
\noindent
\textbf{Image classification.}
The first self-paced dictionary learning method for image classification was proposed by Tang et al.~\citeyearpar{tang2012self}. They employ an easy-to-hard approach that introduces information about the complexity of the samples into the learning procedure. The easy examples are automatically chosen at each iteration, using the learned dictionary from the previous iteration, with more difficult samples being gradually introduced at each step. To enhance the training domain, the number of chosen samples in each iteration is increased using an adaptive threshold function.

Li and Gong~\citeyearpar{li2017self} apply the self-paced learning methodology to convolutional neural networks. The examples are learned from easy to complex, taking into consideration the loss of the model. In order to ensure this schedule, the authors include the self-paced optimization into the learning objective of the CNN, learning both the network parameters and the latent weight variable.

Ren et al.~\citeyearpar{ren2017robust} introduce an SPL model of robust softmax regression for multi-class classification. Their approach computes the complexity of each sample, based on the value of the loss, assigning soft weights according to which the examples are used in the classification problem. Although this method helps to remove the outliers, it can bias the training towards the classes with instances more sensitive to the loss. The authors address this problem by assigning weights and selecting examples locally from each class, using two novel SPL regularization terms.

Zhang et al.~\citeyearpar{zhang2021flexmatch} employ a self-paced learning technique to improve the performance of image classification models in a semi-supervised context. While the traditional approach for selecting the pseudo-labels is to filter them using a predetermined threshold, the authors propose changing the value of the threshold for each class, at every step. Thus, they suggest that the model performs better for a class if many instances of that category are selected when considering a certain threshold. Otherwise, the class threshold is lowered, allowing more examples from the category to be visited. Beside creating a curriculum schedule, the flexible threshold automatically balances the data selection process, ensuring the diversity.

Cascante-Bonilla et al.~\citeyearpar{cascante2020curriculum} propose a curriculum labeling approach that enhances the process of selecting the right pseudo-labels using a curriculum based on Extreme Value Theory. They use percentile scores to decide how many easy samples to add to the training, instead of fixing or manually tuning the thresholds. The difficulty of the pseudo-labeled examples is determined by taking into consideration their loss. Furthermore, to prevent accumulating errors produced at the beginning of the fine-tuning process, the authors allow previous pseudo-annotated samples to enter or leave the new training set. 

Morerio et al.~\citeyearpar{Morerio-ICCV-2017} propose a new regularization technique called curriculum dropout. They show that the standard approach using a fixed dropout probability during training is suboptimal and propose a time scheduling for the probability of retaining neurons in the network. By doing this, the authors increase the difficulty of the optimization problem, generating an easy-to-hard methodology that matches the idea of curriculum learning. They show the superiority of this method over the standard dropout approach by conducting extensive image classification experiments. 

Dogan et al.~\citeyearpar{dogan2020label} propose a label similarity curriculum approach for image classification. Instead of using the actual labels for training the classifier, they use a probability distribution over classes, in the early stages of the learning process. Then, as the training advances, the labels are turned back into the standard one-hot encoding. The intuition is that, at the beginning of the training, it is natural for the model to misclassify similar classes, so that the algorithm only penalizes big mistakes. The authors claim that the similarity between classes can be computed with a predefined metric, suggesting the use of the cosine similarity between embeddings for classes defined by natural language words.

Guo et al.~\citeyearpar{guo2018curriculumnet} propose a curriculum approach for training deep neural networks on large-scale weakly-supervised web images which contain large amounts of noisy labels. Their framework contains three stages: the initial feature generation in which the network is trained for a few iterations on the whole data set, the curriculum design, and the actual curriculum learning step where the samples are presented from easy to hard. They build the curriculum in an unsupervised way, measuring the difficulty with a clustering algorithm based on density.

Choi et al.~\citeyearpar{Choi2019PseudoLabelingCF} apply a similar procedure to generate the curriculum, using clustering based on density, where examples with high density are presented earlier during training than the low-density samples. Their pseudo-labeling curriculum for cross-domain tasks can alleviate the problem of false pseudo-labels. Thus, the network progressively improves the generated pseudo-labels that can be used in the later phases of training.

Shu et al.~\citeyearpar{shu2019transferable} propose a transferable curriculum method for weakly-supervised domain adaptation tasks. The curriculum selects the source samples which are noiseless and transferable, thus are good candidates for training. The framework splits the task into two sub-problems: learning with transferable curriculum, which guides the model from easy to hard and from transferable to untransferable, and constructing the transferable curriculum to quantify the transferability of source examples based on their contributions to the target task. A domain discriminator is trained in a curriculum fashion, which enables measuring the transferability of each sample.

\begin{sloppypar}
Yang et al.~\citeyearpar{YangBLS20} introduce a curriculum procedure for selecting the training samples that maximize the performance of a multi-source unsupervised domain adaptation method. They build the method on top of a domain-adversarial strategy, employing a domain discriminator to separate source and target examples. Then, their framework creates the curriculum by taking into consideration the loss of the discriminator on the source samples, i.e.,~examples with a higher loss are closer to the target distribution and should be selected earlier. The components are trained in an adversarial fashion, improving each other at every step.
\end{sloppypar}

Weinshall and Cohen \citeyearpar{Weinshall2018CurriculumLB} elaborate an extensive investigation of the behavior of curriculum convolutional models with regard to the difficulty of the samples and the task. They estimate the difficulty in a transfer learning fashion, taking into consideration the confidence of a different pre-trained network. They conduct classification experiments under different task difficulty conditions and different scheduling conditions, showing that curriculum learning increases the rate of convergence in the early phases of training.

Hacohen and Weinshall~\citeyearpar{Hacohen2019OnTP} also conduct an extensive analysis of curriculum learning, experimenting on multiple settings for image classification. They model the easy-to-hard procedure in two ways. First, they train a teacher network and use the confidence of its predictions as the scoring function for each image. Second, they train the network conventionally, then compute the confidence score for each image to define a scoring function with which they retrain the model from scratch in a curriculum way. Furthermore, they test multiple pacing functions to determine the impact of the curriculum schedule on the final results.

\begin{sloppypar}
A different approach is taken by Cheng et al. \citeyearpar{cheng2019local}, who replace the easy-to-hard formulation of standard CL with a local-to-global training strategy. The main idea is to first train a model on examples from a certain class, then gradually add more clusters to the training set. Each training round completes when the model converges. The group on which the training commences is randomly selected, while for choosing the next clusters, three different selection criteria are employed. The first one randomly picks the new group and the other two sample the most similar or dissimilar clusters to the groups already selected. Empirical results show that the selection criterion does not impact the superior results of the proposed framework.
\end{sloppypar}

Pentina et al.~\citeyearpar{pentina2015curriculum} introduce CL for multiple tasks to determine the optimal order for learning the tasks to maximize the final performance. As the authors suggest, although sharing information between multiple tasks boosts the performance of learning models, in a realistic scenario, strong relationships can be identified only between a limited number of tasks. This is why a possible optimization is to transfer knowledge only between the most related tasks. Their approach processes multiple tasks in a sequence, sharing knowledge between subsequent tasks. They determine the curriculum by finding the right task order to maximize the overall expected classification performance.

\begin{sloppypar}
Yu et al.~\citeyearpar{yu2020multi} introduce a multi-task curriculum approach for solving the open-set semi-supervised learning task, where out-of-distribution samples appear in unlabeled data. On the one hand, they compute the out-of-distribution score automatically, training the network to estimate the probability of an example of being out-of-distribution. On the other hand, they use easy in-distribution examples from the unlabeled data to train the network to classify in-distribution instances using a semi-supervised approach. Furthermore, to make the process more robust, they employ a joint operation, updating the network parameters and the scores alternately.
\end{sloppypar}

Guo et al.~\citeyearpar{guo2020breaking} tackle the task of automatically finding effective architectures using a curriculum procedure. They start searching for a good architecture in a small space, then gradually enlarge the space in a multistage approach. The key idea is to exploit the previously learned knowledge: once a fine architecture has been found in the smaller space, a larger, better, candidate subspace that shares common information with the previous space can be discovered.

Gong et al.~\citeyearpar{Gong-TIP-2016} tackle the semi-supervised image classification task in a curriculum fashion, using multiple teachers to assess the difficulty of unlabeled images, according to their reliability and discriminability. The consensus between teachers determines the difficulty score of each example. The curriculum procedure constructed by presenting examples from easy to hard provides superior results than regular approaches.

Taking a different approach than Gong et al.~\citeyearpar{Gong-TIP-2016}, Jiang et al.~\citeyearpar{jiang2018mentornet} use a teacher-student architecture to generate the easy-to-hard curriculum. Instead of assessing the difficulty using multiple teachers, their architecture consists of only two networks: MentorNet and StudentNet. MentorNet learns a data-driven curriculum dynamically with StudentNet and guides the student network to learn from the samples which are probably correctly classified. The teacher network can be trained to approximate the predefined curriculum as well as to find a new curriculum in the data, while taking into consideration the student's feedback.

Kim and Choi~\citeyearpar{kim2018screenernet} also propose a teacher-student curriculum methodology, where the teacher determines the optimal weights to maximize the student's learning progress. The two networks are jointly trained in a self-paced fashion, taking into consideration the student's loss. The method uses a local optimal policy that predicts the weights of training samples at the current iteration, giving higher weights to the samples producing higher errors in the main network. The authors claim that their method is different from the MentorNet introduced by Jiang et al.~\citeyearpar{jiang2018mentornet}. MentorNet is pre-trained on other data sets than the data set of interest, while the teacher proposed by Kim and Choi \citeyearpar{kim2018screenernet} only sees examples from the data set it is applied on.

CL has also been investigated in incremental learning settings. Incremental or continual learning refers to the task in which multiple subsequent training phases share only a partial set of the target classes. This is considered a very challenging task due to the tendency of neural networks to forget what was learned in the preceding training phases, also called \textit{catastrophic forgetting} \citep{McCloskey-PLM-1989}. In \citep{kim2019incremental}, the authors use CL as a side task to remove hard samples from mini-batches, in what they call \emph{DropOut Sampling}. The goal is to avoid optimizing on potentially incorrect knowledge.

Ganesh and Corso~\citeyearpar{ganesh2020rethinking} propose a two-stage approach in which class incremental training is performed first, using a label-wise curriculum. In a second phase, the loss function is optimized through adaptive compensation on misclassified samples. This approach is not entirely classifiable as incremental learning since all past data are available at every step of the training. However, the curriculum is applied in a label-wise manner, adding a certain label to the training starting from the easiest down to the hardest.

Pi et al.~\citeyearpar{pi2016self} combine boosting with self-paced learning in order to build the self-paced boost learning methodology for classification. Although boosting may seem the exact opposite of SPL, focusing on the misclassified (harder) examples, the authors suggest that the two approaches are complementary and may benefit from each other. While boosting reflects the local patterns, being more sensitive to noise, SPL explores the data more smoothly and robustly. The easy-to-hard self-paced schedule is applied to boosting optimization, making the framework focus on the reliable examples which have not yet been learned sufficiently.

Zhou and Bilmes \citeyearpar{zhou2018minimax} also adopt a learning strategy based on selecting the most difficult examples, but they enhance it by taking into consideration the diversity at each step. They argue that diversity is more important during the early phases of training when only a few samples are selected. They also claim that by selecting the hardest samples instead of the easiest, the framework avoids successive rounds selecting similar sets of examples. The authors employ an arbitrary non-monotone submodular function to measure diversity while using the loss function to compute the difficulty.

Zhou et al.~\citeyearpar{zhou2020curriculum} introduce a new measure, dynamic instance hardness (DIH), to capture the difficulty of examples during training. They use three types of instantaneous hardness to compute DIH: the loss, the loss change, and the prediction flip between two consecutive time steps. The proposed approach is not a standard easy-to-hard procedure. Instead, the authors suggest that training should focus on the samples that have historically been hard since the model does not perform or generalize well on them. Hence, in the first training steps, the model will warm-up by sampling examples randomly. Then, it will take into consideration the DIH and select the most difficult examples, which will also gradually become easier.

Ren et al.~\citeyearpar{ren18a} introduce a re-weighting meta-learning scheme that learns to assign weights to training examples based on their gradient directions. The authors claim that the two contradicting loss-based approaches, SPL and HEM, should be employed in different scenarios. Therefore, in noisy label problems, it is better to emphasize the smaller losses, while in class imbalance problems, algorithms based on determining the hard examples should perform better. To address this issue, they guide the training using a small unbiased validation set. Thus, the new weights are determined by performing a meta-gradient descent step on each mini-batch to minimize the loss on the clean validation set.

Tang and Huang~\citeyearpar{tang2019self} combine active learning and SPL, creating an algorithm that introduces the right training examples at the right moment. Active learning selects the samples having the highest potential of improving the model. Still, those examples can be easy or hard, and including a difficult sample too early during training might limit its benefit. To address this issue, the authors propose to jointly consider the potential value and the easiness of instances. In this way, the selected examples will be both informative and easy enough to be utilized by the current model. This is achieved by applying two weights for each unlabeled instance, one that estimates the potential value on improving the model and another that captures the difficulty of the example.

Chang et al.~\citeyearpar{chang2017active} use an active learning approach based on sample uncertainty to improve learning accuracy in multiple tasks. The authors claim that SPL and HEM might work well in different scenarios, but sorting the examples based on the level of uncertainty might provide a universal solution to improve the performance of the model. The main idea is that the examples predicted correctly with high confidence may be too easy to contain useful information for improving that model further, while the examples which are constantly predicted incorrectly may be too difficult and degrade the model. Hence, the authors focus on the uncertain samples, modeling the uncertainty in two ways: using the variance of the predicted probability of the correct class during learning and the closeness of the correct class probability to the decision threshold.

\vspace{0.2cm}
\noindent
\textbf{Object detection and localization.}
Shi and Ferrari~\citeyearpar{shi2016weakly} employ a standard curriculum approach, using size estimates to assess the difficulty of images in a weakly-supervised object localization task. They assume that images with bigger objects are easier and build a regressor that can estimate the size of objects. They use a batching approach, splitting the set into $n$ shards based on the difficulty, then beginning the training process on the easiest batch, and gradually adding the more difficult groups.

Li et al.~\citeyearpar{Li-BMVC-2017} use curriculum learning for weakly-supervised object detection. Over the standard detector, they add a segmentation network that helps to detect the full objects. Then, they employ an easy-to-hard approach for the re-localization and retraining steps, by computing the consistency between the outputs from the detector and the segmentation model, using intersection over reunion (IoU). The examples which have the IoU value greater than a preselected threshold are easier, thus they will be used in the first steps of the algorithm.

Wang et al.~\citeyearpar{Wang-ICPR-2018} introduce an easy-to-hard curriculum approach for weakly and semi-supervised object detection. The framework consists of two stages: first, the detector is trained using the fully annotated data, then it is fine-tuned in a curriculum fashion on the weakly annotated images. The easiness is determined by training an SVM on a subset of the fully annotated data and measuring the mean average precision per image (mAPI): an example is easy if its mAPI is greater than 0.9, difficult if its mAPI is lower than 0.1, and normal otherwise.

\begin{sloppypar}
Sangineto et al.~\citeyearpar{sangineto2018self} use SPL for weakly-supervised object detection. They use multiple rounds of SPL in which they progressively enhance the training set with pseudo-labels that are easy to predict. In this methodology, the easiness is defined by the reliability (confidence) of each pseudo-bounding box. Furthermore, the authors introduce self-paced learning at the class level, using the competition between multiple classifiers to estimate the difficulty of each class.
\end{sloppypar}

Zhang et al.~\citeyearpar{Zhang-IJCV-2019} propose a collaborative SPCL framework for weakly-supervised object detection. Compared to Jiang et al.~\citeyearpar{Jiang-AAAI-2015}, the collaborative SPCL approach works in a setting where the data set is not fully annotated, corroborating the confidence at the image level with the confidence at the instance level. The image-level predefined curriculum is generated by counting the number of labels for each image, considering that examples with multiple object categories have a larger ambiguity, thus being more difficult. To compute the instance-level difficulty, the authors employ a mask-out strategy, using an AlexNet-like architecture pre-trained on ImageNet to determine which instances are more likely to contain objects from certain categories.  Starting from this prior knowledge, the self-pacing regularizers use both the instance-level and image-level sample confidence to build the curriculum.

Soviany et al.~\citeyearpar{soviany2019curriculum} apply a self-paced curriculum learning approach for unsupervised cross-domain object detection. They propose a multi-stage framework in order to better transfer the knowledge from the source domain to the target domain. In this first stage, they employ a cycle-consistency GAN \citep{Zhu-ICCV-2017} to translate images from source to target, thus generating fully annotated data with a similar aspect to the target domain. In the next step, they train an object detector on the original source images and on the translated examples, then they generate pseudo-labels to fine-tune the model on target data. During the fine-tuning process, an easy-to-hard curriculum is applied to select highly confident examples. The curriculum is based on a difficulty metric given by the number of detected objects divided by their size.

Shrivastava et al.~\citeyearpar{shrivastava2016training} introduce the online hard example mining (OHEM) algorithm for object detection, which selects training examples according to the current loss of each sample. Although the idea is similar to SPL, the methodology is the exact opposite. Instead of focusing on the easier examples, OHEM favors diverse high-loss instances. For each training image, the algorithm extracts the regions of interest (RoIs) and performs a forward step to compute the losses, then sorts the RoIs according to the loss. It selects only the training samples for which the current model performs badly, having high loss values.

\vspace{0.2cm}
\begin{sloppypar}
\noindent
\textbf{Object segmentation.}
Kumar et al.~\citeyearpar{kumar2011learning} apply a similar procedure to the original formulation of SPL proposed in \citep{Kumar-NIPS-2010} in order to determine class-specific segmentation masks from diverse data. The strategy is applied to different kinds of data, e.g.,~segmentation masks with generic foreground or background classes, to identify the specific classes and bounding box annotations and to determine the segmentation masks. In their experiments, the authors use a latent structural SVM with SPL based on the likelihood to predict the correct output.
\end{sloppypar}

Zhang et al.~\citeyearpar{zhang2017curriculum} apply a curriculum strategy to the semantic segmentation task for the domain adaptation scenario. They use simpler, intermediate tasks to determine certain properties of the target domain which lead to improved results on the main segmentation task. This strategy is different from the previous examples because it does not only order the training samples from easy to hard, but it shows that solving some simpler tasks provides information that allows the model to obtain better results on the main problem.

Sakaridis et al.~\citeyearpar{sakaridis2019guided} use a curriculum approach for adapting semantic segmentation models from daytime to nighttime, in an unsupervised fashion, starting from a similar intuition as Zhang et al.~\citeyearpar{zhang2017curriculum}. Their main idea is that models perform better when more light is available. Thus, the easy-to-hard curriculum is treated as daytime to nighttime transfer, by training on multiple intermediate light phases, such as twilight. Two kinds of data are used to present the progressively darker times of the day: labeled synthetic images and unlabeled real images.

As Sakaridis et al.~\citeyearpar{sakaridis2019guided}, Dai et al.~\citeyearpar{dai2020curriculum} apply a methodology for adapting semantic segmentation models from fogless images to a dense fog domain. They use the fog density to rank images from easy to hard and, at each step, they adapt the current model to the next (more difficult) domain, until reaching the final, hardest, target domain. The intermediate domains contain both real and artificial data which make the data sets richer. To estimate the difficulty of the examples, i.e., the level of fog, the authors build a regression model over artificially generated images with fixed levels of fog.

\begin{sloppypar}
Feng et al.~\citeyearpar{feng2020semi} propose a curriculum self-training approach for semi-supervised semantic segmentation. The fine-tuning process selects only the most confident $\alpha$ pseudo-labels from each class. The easy-to-hard curriculum is enforced by applying multiple self-training rounds and by increasing the value of $\alpha$ at each round. Since the value of $\alpha$ decides how many pseudo-labels to be activated, the higher value allows lower confidence (more difficult) labels to be used during the later phases of learning.
\end{sloppypar}

Qin et al.~\citeyearpar{qin2020balanced} introduce a curriculum approach that balances the loss value with respect to the difficulty of the samples. The easy-to-hard strategy is constructed by taking into consideration the classification loss of the examples: samples with low classification loss are far away from the decision boundary and, thus, easier. They use a teacher-student approach, jointly training the teacher and the student networks. The difficulty is determined by the predictions of the teacher network, being subsequently employed to guide the training of the student. The curriculum methodology is applied to the student model, by decreasing the loss of difficult examples and increasing the loss of easy examples. 

\vspace{0.2cm}
\noindent
\textbf{Face recognition.}
Buyuktas et al.~\citeyearpar{buyuktas2020curriculum} suggest a classic curriculum batching strategy for face recognition. They present the training data from easy to hard, computing the difficulty using the head pose as a measure of difficulty, with images containing upright frontal faces being the easiest to recognize. The authors obtain the angle of the head pose using features like yaw, pitch and roll angles. Their experiments show that the CL approach improves the random baseline by a significant margin. 

\begin{sloppypar}
Lin et al.~\citeyearpar{lin2017active} combine the opposing active learning (AL) and self-paced learning methodologies to build a ``cost-less-earn-more'' model for face identification. After the model is trained on a limited number of examples, the SPL and AL approaches are employed to generate more data. On the one hand, easy (high-confidence) examples are used to obtain reliable pseudo-labels on which the training proceeds. On the other hand, the low-confidence samples, which are the most informative in the AL scenario, are selected, using an uncertainty-based strategy, to be annotated with human supervision. Furthermore, two alternative types of curriculum constraints, which can be dynamically changed as the training progresses, are applied to guide the training.
\end{sloppypar}

Huang et al.~\citeyearpar{huang2020curricularface} introduce a difficulty-based method for face recognition. Unlike traditional CL and SPL methods, which gradually enhance the training set with more difficult data, the authors design a loss function that guides the learning through an easy-then-hard strategy inspired by HEM. Concretely, this new loss emphasizes easy examples at the beginning of the training and hard samples in the later stages of the learning process. In their framework, the samples are randomly selected in each mini-batch, and the curriculum is established adaptively using HEM. Furthermore, the impact of easy and hard samples is dynamic and can be adjusted in different training stages.

\vspace{0.2cm}
\noindent
\textbf{Image generation and translation.}
Soviany et al. \citeyearpar{Soviany-WACV-2020} apply curriculum learning in their image generation and image translation experiments using generative adversarial networks (GANs). They use the image difficulty estimator from \citep{Ionescu-CVPR-2016} to rank the images from easy to hard and apply three different methods (batching, sampling and weighing) to determine how the difficulty of real data examples impacts the final results. The last two methods are based on an easiness score which converges to the value $1$ as the training advances. The easiness score is either used to sample examples (in curriculum by sampling) or integrated into the discriminator loss function (in curriculum by weighting).

While Soviany et al.~\citeyearpar{Soviany-WACV-2020} use a standard data-level curriculum, Karras et al.~\citeyearpar{Karras-ICLR-2018} propose a model-based method for improving the quality of GANs. By gradually increasing the size of the generator and discriminator networks, the training starts with low-resolution images, then the resolution is progressively increased by adding new layers to the model. Thus, the implicit curriculum learning is determined by gradually increasing the network's capacity, allowing the model to focus on the large-scale structure of the image distribution at first, then concentrate on the finer details later.

For improving the performance of GANs, Doan et al.~\citeyearpar{Doan-AAAI-2019} propose training a single generator on a target data set using curriculum over multiple discriminators. They do not employ an easy-to-hard strategy, but, through the curriculum, they attempt to optimally weight the feedback received by the generator according to the status of each discriminator. Hence, at each step, the generator is trained using the combination of discriminators providing the best learning information. The progress of the generator is used to provide meaningful feedback for learning efficient mixtures of discriminators.

\vspace{0.2cm}
\noindent
\textbf{Video processing.}
Tang et al.~\citeyearpar{tang2012shifting} adopt an SPL strategy for unsupervised adaptation of object detectors from image to video. The training procedure is simple: an object detector is first trained on labeled image data to detect the presence of a certain class. The detector is then applied to unlabeled videos to detect the top negative and positive examples, using track-based scoring. In the self-paced steps, the easiest samples from the video domain, together with the images from the source domain are used to retrain the model. As the training progresses, more data samples from the video domain are added, while the most difficult samples from the image domain are removed. The easiness is seen as a function of the loss, the examples with higher losses being labeled as more difficult.

Supancic and Ramanan~\citeyearpar{supancic2013self} use an SPL methodology for addressing the problem of long-term object tracking. The framework has three different stages. In the first stage, a detector is trained given a set of positive and negative examples. In the second stage, the model performs tracking using the previously learned detector and, in the final stage, the tracker selects a subset of frames from which to re-learn the detector for the next iteration. To determine the easy samples, the model finds the frames that produce the lowest SVM objective when added to the training set.

\begin{sloppypar}
Jiang et al.~\citeyearpar{jiang2014easy} introduce a self-paced re-ranking approach for multimedia retrieval. As the authors note, traditional re-ranking systems assign either binary weights, so the top videos have the same importance, or predefined weights which might not capture the actual importance in the latent space. To solve this issue, they suggest assigning the weights adaptively using an SPL approach. The models learn gradually, from easy to hard, recomputing the easiness scores based on the actual training progress, while also updating the model's weights. 
\end{sloppypar}

Furthermore, Jiang et al.~\citeyearpar{jiang2014self} introduce a self-paced learning with diversity methodology to extend the standard easy-to-hard approaches. The intuition behind it correlates with the way people learn: a student should see easy samples first, but the examples should be diverse enough to understand the concept. Furthermore, the authors show that an automatic model which uses SPL and has been initialized on images from a certain group will be biased towards easy examples from the same group, ignoring the other easy samples and leading to overfitting. In order to select easy and diverse examples, the authors add a new term to the classic SPL regularization, namely the negative $l_{2,1}$-norm, which favors selecting samples from multiple groups. Their event detection and action recognition results outperform the results of the standard SPL approach.

Liang et al.~\citeyearpar{liang2016learning} propose a self-paced curriculum learning approach for training detectors that can recognize concepts in videos. They apply prior knowledge to guide the training, while also allowing model updates based on the current learning progress. To generate the curriculum component, the authors take into consideration the term frequency in the video metadata. Moreover, to improve the standard CL and SPL approaches, they introduce partial-order curriculum and dropout, which can enhance the model's results when using noisy data. The partial-order curriculum leverages the incomplete prior information, alleviating the problem of determining a learning sequence for every pair of samples, when not enough examples are available. The dropout component provides a way of combining different sample subsets at different learning stages, thus preventing overfitting to noisy labels. 

Zhang et al.~\citeyearpar{zhang2017spftn} present a deep learning approach for object segmentation in weakly labeled videos. By employing a self-paced fine-tuning network, they manage to obtain good results by using positive videos only. The self-paced regularizer used to guide the training has two components: the traditional SPL function which captures the sample easiness and a group curriculum term. The curriculum uses predefined learning priorities to favor training samples from certain groups.

\vspace{0.2cm}
\begin{sloppypar}
\noindent
\textbf{Other tasks.}
Guy et al.~\citeyearpar{Gui-FG-2017} extend curriculum learning to the facial expression recognition task. They consider the idea of expression intensity to measure the difficulty of a sample: the more intense the expression is (a large smile for happiness, for example), the easier it is to recognize. They employ an easy-to-hard batching strategy which leads to better generalization for emotion recognition from facial expressions.
\end{sloppypar}

\begin{sloppypar}
Sarafianos et al.~\citeyearpar{sarafianos2017curriculum} combine the advantages of multi-task and curriculum learning for solving a visual attribute classification problem. In their framework, they group the tasks into strongly-correlated tasks and weakly-correlated tasks. In the next step, the training commences following a curriculum procedure, starting on the strongly-correlated attributes, and then transferring the knowledge to the weakly-correlated group. In each group, the learning process follows a multitask classification setup.
\end{sloppypar}

Wang et al.~\citeyearpar{Wang_2019_ICCV} introduce the dynamic curriculum learning framework for imbalanced data learning that is composed of two types of curriculum schedulers. On the one hand, the sampling scheduler detects the most meaningful samples in each batch to guide the training from imbalanced to balanced and from easy to hard. On the other hand, the loss scheduler adjusts the learning weights between the classification loss and the metric learning loss. An example is considered easy if it is correctly predicted. The evaluation of two attribute analysis data sets shows the superiority of the approach over conventional training.

Lee and Grauman~\citeyearpar{lee2011learning} propose an SPL approach for visual category discovery. They do not use a predefined teacher to guide the training in a pure curriculum way. Instead, they are constraining the model to automatically select the examples which are easy enough at a certain time during the learning process. To define easiness, the authors consider two concepts: objectness and context-awareness. The algorithm discovers objects, from one category at a time, in the order of the predicted easiness. After each discovery, the difficulty score is updated, and the criterion for the next stage is relaxed.

\begin{sloppypar}
Zhang et al.~\citeyearpar{zhang2015self} use a self-paced methodology for multiple-instance learning (MIL) in co-saliency detection. As the authors suggest, MIL is a natural method for solving co-saliency detection, exploring both the contrast between co-salient objects and contexts and the consistency of co-salient objects in multiple images. Furthermore, to obtain reliable instance annotations and instance detections, they combine MIL with easy-to-hard SPL. The framework focuses on co-salient image regions from high-confidence instances first, gradually switching to more complex examples. Moreover, a term for computing the diversity, which penalizes examples selected from the same group, is added to the regularizer. Experimental results show the importance of both easy-to-hard strategy and diverse sampling.
\end{sloppypar}

\begin{sloppypar}
Xu et al.~\citeyearpar{xu2015multi} introduce a multi-view self-paced learning method for clustering that applies the easy-to-hard methodology simultaneously at the sample level and the view level. They apply the difficulty using a probabilistic smoothed weighting scheme, instead of standard binary labels. Whereas SPL regularization has already been employed on examples, the concept of computing the difficulty of the view is new. As the authors suggest, a multi-view example can be more easily distinguishable in one view than in the others, because the views present orthogonal perspectives, with different physical meanings.
\end{sloppypar}

Zhou et al.~\citeyearpar{zhou2018deep} propose a self-paced approach for alleviating the problem of noise in a person re-identification task. Their algorithm contains two main components: a self-paced constraint and a symmetric regularization. The easy-to-hard self-paced methodology is enforced using a soft polynomial regularization term taking into consideration the loss and the age of the model. The symmetric regularizer is applied to minimize the intra-class distance while also maximizing the inter-class distance for each training sample.

Duan et al.~\citeyearpar{DBLP:conf/eccv/DuanZ0YNG20} introduce a curriculum approach for learning a continuous Signed Distance Function on shapes. They develop their easy-to-hard strategy based on two criteria: surface accuracy and sample difficulty. The surface accuracy is computed using stringency in supervising, with ground truth, while the sample difficulty considers points with incorrect sign estimations as hard. Their method is built to first learn to reconstruct coarse shapes, then focus on more complex local details.

Ghasedi et al.~\citeyearpar{Ghasedi-CVPR-2019} propose a clustering framework consisting of three networks: a discriminator, a generator and a clusterer. They use an adversarial approach to synthesize realistic samples, then learn the inverse mapping of the real examples to the discriminative embedding space. To ensure an easy-to-hard training protocol they employ a self-paced learning methodology, while also taking into consideration the diversity of the samples. Besides the standard computation of the difficulty based on the current loss, they use a lasso regularization to ensure diversity and prevent learning only from the easy examples in certain clusters.

\subsection{Natural language processing}

Multiple works show that many of the curriculum strategies which have been proven to work well on vision tasks can also be employed in various NLP problems. Usual metrics for the vanilla curriculum approach involve domain-specific features based on linguistic information, such the length of the sentences, the number of coordinating conjunctions or word rarity~\citep{spitkovsky2009babysteps,kocmi2017curriculum,zhang2018empirical,platanios2019competence}.

\vspace{0.2cm}
\noindent
\textbf{Machine translation.}
Kocmi and Bojar~\citeyearpar{kocmi2017curriculum} employ a standard easy-to-hard curriculum batching strategy for machine translation. They employ the length of the sentences, the number of coordinating conjunction and the word frequency to assess the difficulty. They constrain the model so that each example is only seen once during an epoch.

Platanios et al.~\citeyearpar{platanios2019competence} propose a similar continuous curriculum learning framework for neural machine translation. They also use the sentence length and the word rarity to compute the difficulty of the examples. During training, they determine the competence of the model, i.e., the learning progress, and sample examples that have the difficulty score lower than the current competence.

Zhang et al.~\citeyearpar{zhang2018empirical} perform an extensive analysis of curriculum learning on a German-English translation task. They measure the difficulty of the examples in two ways: using an auxiliary translation model or taking into consideration linguistic information (term frequency, sentence length). They use a non-deterministic sampling procedure that assigns weights to shards of data, by taking into consideration the difficulty of the examples and the training progress. Their experiments show that it is possible to improve the convergence time without losing translation quality. The results also highlight the importance of finding the right difficulty criterion and curriculum schedule.

\begin{sloppypar}
Guo et al.~\citeyearpar{guo2020fine} use curriculum learning for non-autoregressive machine translation. The main idea comes from the fact that non-autoregressive translation (NAT) is a more difficult task than the standard autoregressive translation (AT), although they share the same model configuration. This is why the authors tackle this problem as a fine-tuning from AT to NAT, employing two kinds of curriculum: a curriculum for the decoder input and a curriculum for the attention mask. This method differs from standard curriculum approaches because the easy-to-hard strategy is applied to the training mechanisms.
\end{sloppypar}

\begin{sloppypar}
Liu et al.~\citeyearpar{liu2020task} propose another curriculum approach for training a NAT model starting from AT. They introduce semi-autoregressive translation (SAT) as intermediate tasks that are tuned using a hyperparameter $k$, which defines an SAT task with different degrees of parallelism. The easy-to-hard curriculum schedule is built by gradually incrementing the value of $k$ from 1 to the length of the target sentence. The authors claim that their method differs from the one of Guo et al.~\citeyearpar{guo2020fine} because they do not use hand-crafted training strategies, but intermediate tasks, showing strong empirical motivation.
\end{sloppypar}

Zhang et al.~\citeyearpar{zhang2019curriculum} use curriculum learning for machine translation in a domain adaptation setting. The difficulty of the examples is computed as the distance (similarity) to the source domain, so that ``more similar examples are seen earlier and more frequently during training'' \citep{zhang2019curriculum}. The data samples are grouped in difficulty batches, and the training commences on the easiest shard. As the training progresses, the more difficult batches become available, until reaching the full data set.

Wang et al.~\citeyearpar{wang-etal-2019-dynamically} introduce a co-curriculum strategy for neural machine translation, combining two levels of heuristics to generate a domain curriculum and a denoising curriculum. The domain curriculum gradually removes fewer in-domain samples, optimizing towards a specific domain, while the denoising curriculum gradually discards noisy examples to improve the overall performance of the model. They combine the two curricula with a cascading approach, gradually discarding examples that do not fit both requirements. Furthermore, the authors employ optimization to the co-curriculum, iteratively improving the denoising selection without losing quality on the domain selection.

\begin{sloppypar}
Wang et al.~\citeyearpar{wang2020learning} introduce a multi-domain curriculum approach for neural machine translation. While the standard curriculum procedure discards the least useful examples according to a single domain, their weighting scheme takes into consideration all domains when updating the weights. The intuition is that a training example that improves the model for all domains produces gradient updates leading the model towards a common direction in all domains. Since this is difficult to achieve by selecting a single example, they propose to work in a data batch on average, building a trade-off between regularization and domain adaptation.
\end{sloppypar}

\begin{sloppypar}
Zhan et al.~\citeyearpar{zhan2021meta} propose a meta-curriculum learning approach for addressing the problem of neural machine translation in a cross-domain setting. They build their easy-to-hard curriculum starting with the common elements of the domains, then progressively addressing more specific elements. To compute the commonalities and the individualities, they apply pre-trained neural language models. Their experimental results show that adding curriculum learning over meta-learning for cross-domain neural machine translations improves the performance on domains previously seen during training, but also on the unseen domains.
\end{sloppypar}

Kumar et al.~\citeyearpar{kumar2019reinforcement} use a meta-learning curriculum approach for neural machine translation. They employ a noise estimator to predict the difficulty of the examples and split the training set into multiple bins according to their difficulty. The main difference from the standard CL is the learning policy which does not automatically proceed from easy to hard. Instead, the authors employ a reinforcement learning approach to determine the right batch to sample at each step, in a single training run. They model the reward function to measure the delta improvement with respect to the average reward recently received, lowering the impact of the tasks selected at the beginning of the training. 

Liu et al.~\citeyearpar{liu-etal-2020-norm} propose a norm-based curriculum learning method for improving the efficiency of training a neural machine translation system. They use the norm of a word embedding to measure the difficulty of the sentence, the competence of the model, and the weight of the sentence. The authors show that the norms of the word vectors can capture both model-based and linguistic features, with most of the frequent or rare words having vectors with small norms. Furthermore, the competence component allows the model to automatically adjust the curriculum during training. Then, to enhance learning even more, the difficulty levels of the sentences are transformed into weights and added to the objective function.

\begin{sloppypar}
Ruiter et al.~\citeyearpar{ruiter2020self} analyze the behavior of self-supervised neural machine translation systems that jointly learn to select the right training data and to perform translation. In this framework, the two processes are designed in such a fashion that they enable and enhance each other. The authors show that the sampling choices made by these models generate an implicit curriculum that matches the principles of CL: samples are self-selected based on increasing complexity and task-relevance, while also performing a denoising curriculum.
\end{sloppypar}

Zhao et al.~\citeyearpar{ZhaoWNW20} introduce a method for generating the right curriculum for neural machine translation. The authors claim that this task highly relies on large quantities of data that are hard to acquire. Hence, they suggest re-selecting influential data samples from the original training set. To discover which examples from the existing data set may further improve the model, the re-selection is designed as a reinforcement learning problem. The state is represented by the features of randomly selected training instances, the action is selecting one of the samples, and the reward is the perplexity difference on a validation set, with the final goal of finding the policy that maximizes the reward.

\begin{sloppypar}
Zhou et al.~\citeyearpar{zhou2020uncertainty} introduce an uncertainty-based curriculum batching approach for neural machine translation. They propose using uncertainty at the data level, for establishing the easy-to-hard ordering, and the model level, to decide the right moment to enhance the training set with more difficult samples. To measure the difficulty of the examples, they start from the intuition that samples with higher cross-entropy and uncertainty are more difficult to translate. Thus, the data uncertainty is measured according to its joint distribution, as it is estimated by a language model pre-trained on the training data. On the other hand, the model's uncertainty is evaluated using the variance of the distribution over a Bayesian neural network.
\end{sloppypar}

\vspace{0.2cm}
\noindent
\textbf{Question answering.}
Sachan and Xing~\citeyearpar{sachan2016easy} propose new heuristics for determining the easiness of examples in an SPL scenario, other than the standard loss function. Aside from the heuristics, the authors highlight the importance of diversity. They measure diversity using the angle between the hyperplanes that the question examples induce in the feature space. Their solution selects a question that is valid according to both criteria, being easy, but also diverse with regards to the previously sampled examples.

Liu et al.~\citeyearpar{liu2018curriculum} introduce a curriculum learning framework for natural answer generation that learns a basic model at first, using simple and low-quality question-answer (QA) pairs. Then, it gradually introduces more complex and higher-quality QA pairs to improve the quality of the generated content. The authors use the term frequency selector and a grammar selector to assess the difficulty of the training examples. The curriculum methodology is ensured using a sampling strategy which gives higher importance to easier examples, in the first iterations, but equalizes it, as the training advances.

Bao et al.~\citeyearpar{bao2020plato} use a two-stage curriculum learning approach for building an open-domain chatbot. In the first, easier stage, a simplified one-to-one mapping modeling is imposed to train a coarse-grained generation model for generating responses in various conversation contexts. The second stage moves to a more difficult task, using a fine-grained generation model and an evaluation model. The most appropriate responses generated by the fine-grained model are selected using the evaluation model, which is trained to estimate the coherence of the responses.

\vspace{0.2cm}
\begin{sloppypar}
\noindent
\textbf{Other tasks.}
The importance of presenting the data in a meaningful order is highlighted by Spitkovsky et al.~\citeyearpar{spitkovsky2009babysteps} in their unsupervised grammar induction experiments. They use the length of a sentence as a difficulty metric, with longer sentences being harder, suggesting two approaches: ``Baby steps'' and ``Less is more''. ``Baby steps'' shows the superiority of an easy-to-hard training strategy by iterating over the data in increasing order of the sentence length (difficulty) and augmenting the training set at each step. ``Less is more'' matches the findings of Bengio et al.~\citeyearpar{Bengio-ICML-2009} that sometimes easier examples can be more informative. Here, the authors improve the state of the art while limiting the sentence length to a maximum of 15.
\end{sloppypar}

\begin{sloppypar}
Zaremba and Sutskever~\citeyearpar{zaremba2014learning} apply curriculum learning to the task of evaluating short code sequences of \emph{length} = $a$ and \emph{nesting} = $b$. They use the two parameters as difficulty metrics to enforce a curriculum methodology. Their first procedure is similar to the one in \citep{Bengio-ICML-2009}, starting with the \emph{length} = 1 and \emph{nesting} = 1, while iteratively increasing the values until reaching $a$ and $b$, respectively. To improve the results of this naive approach, the authors build a mixed technique, where the values for length and nesting are randomly selected from $[1,a]$ and $[1,b]$. The last method is a combination of the previous two approaches. In this way, even though the model still follows an easy-to-hard strategy, it has access to more difficult examples in the early stages of the training.
\end{sloppypar}

Tsvetkov et al.~\citeyearpar{tsvetkov2016learning} introduce Bayesian optimization to optimize curricula for word representation learning. They compute the complexity of each paragraph of text using three groups of features: diversity, simplicity, and prototypicality. Then, they order the training set according to complexity, generating word representations that are used as features in a subsequent NLP task. Bayesian optimization is applied to determine the best features and weights that maximize performance on the chosen NLP task.

Cirik et al.~\citeyearpar{cirik2016visualizing} analyze the effect of curriculum learning on training Long Short-Term Memory (LSTM) networks. For that, they employ two curriculum strategies and two baselines. The first curriculum approach uses an easy-then-hard methodology, while the second one is a batching method which gradually enhances the training set with more difficult samples. As baselines, the authors choose conventional training based on random data shuffling and an option where, at each epoch, all samples are presented to the network, ordered from easy to hard. Furthermore, the authors analyze CL with regard to the model complexity and available resources.

Graves et al.~\citeyearpar{Graves-ICML-2017} tackle the problem of automatically determining the path of a neural network through a curriculum to maximize learning efficiency. They test two related setups. In the multiple tasks setup, the challenge is to achieve high results on all tasks, while in the target task setup, the goal is to maximize the performance on the final task. The authors model the curriculum over $n$ tasks as an $n$-armed bandit, and a syllabus as an adaptive policy seeking to maximize the rewards from the bandit. 

Subramanian et al.~\citeyearpar{subramanian2017adversarial} employ adversarial architectures to generate natural language. They define the curriculum learning paradigm by constraining the generator to produce sequences of gradually increasing lengths as training progresses. Their results show that the curriculum ordering is essential when generating long sequences with an LSTM.

Huang and Du~\citeyearpar{huang2019self} introduce a collaborative curriculum learning framework to reduce the impact of mislabeled data in distantly supervised relation extraction. In the first step, they employ an internal self-attention mechanism between the convolution operations which can enhance the quality of sentence representations obtained from the noisy inputs. Next, a curriculum methodology is applied to two sentence selection models. These models behave as relation extractors, and collaboratively learn and regularize each other. This mimics the learning behavior of two students that compare their different answers. The learning is guided by a curriculum that is generated taking into consideration the conflicts between the two networks or the value of the loss function.

Tay et al.~\citeyearpar{tay-etal-2019-simple} propose a generative curriculum pre-training method for solving the problem of reading comprehension over long narratives. They use an easy-to-hard curriculum approach on top of a pointer-generator model which allows the generation of answers even if they do not exist in the context, thus enhancing the diversity of the data. The authors build the curriculum considering two concepts of difficulty: answerability and understandability. The answerability measures whether an answer exists in the context, while understandability controls the size of the document. 

Xu et al.~\citeyearpar{xu2020curriculum} attempt to improve the standard ``pre-train then fine-tune'' paradigm which is broadly used in natural language understanding, by replacing the traditional training from the fine-tuning stage, with an easy-to-hard curriculum. To assess the difficulty of an example, they measure the performance of multiple instances of the same model, trained on different shards of the data set, except the one containing the example itself. In this way, they obtain an ordering of the samples which they use in a curriculum batching strategy for training the same model.

\begin{sloppypar}
Penha and Hauff~\citeyearpar{penha2019curriculum} investigate curriculum strategies for information retrieval, focusing on conversation response ranking. They use multiple difficulty metrics to rank the examples from easy to hard: information spread, distraction in responses, response heterogeneity, and model confidence. Furthermore, they experiment with multiple methods of selecting the data, using a standard batching approach and other continuous sampling methods.
\end{sloppypar}

Li et al.~\citeyearpar{li2020label} propose a label noise-robust curriculum batching strategy for deep paraphrase identification. They use a combination of two predefined metrics in order to create the easy-to-hard batches. The first metric uses the losses of a model trained for only a few iterations. Starting from the intuition that neural networks learn fast from clean samples and slowly from noisy samples, they design the loss-based noise metric as the mean value of a sequence of losses for training examples in the first epochs. The other criterion is the similarity-based noise metric which computes the similarity between the two sentences using the Jaccard similarity coefficient.

Chang et al.~\citeyearpar{chang2021does} apply curriculum learning for data-to-text generation. They experiment with multiple difficulty metrics and show that measures which consider data and text jointly provide better results than measures which capture only the complexity of data or text. In their curriculum setup, the authors select only the examples which are easy enough, given the competence of the model at the current training step. Their experimental results show that, besides enhancing the quality of the outputs, curriculum learning helps to improve convergence speed.

Gong et al.~\citeyearpar{gong2021density} introduce a dynamic curriculum learning framework for intent detection. They model the difficulty of the training examples using the eigenvectors' density, where a higher density denotes an easier sample. Their dynamic scheduler ensures that, as the training progresses, the number of easy examples is reduced and the number of complex samples is increased. The experimental results show that the proposed method improves both the traditional training baseline and other curriculum learning strategies.

Zhao et al.~\citeyearpar{zhao2021automatic} introduce a curriculum learning methodology for enhancing dialogue policy learning. Their framework involves a teacher-student mechanism which takes into consideration both difficulty and diversity. The authors capture the difficulty using the learning progress of the agent, while penalizing over-repetitions in order to enforce diversity. Furthermore, they introduce three different curriculum scheduling approaches and prove that all of them improve the standard random sampling method.

Jafarpour et al.~\citeyearpar{jafarpour2021active} investigate the benefits of combining active learning and curriculum learning for solving the named entity recognition tasks. They compute the complexity of the examples using multiple linguistic features, including seven novel difficulty metrics. From the perspective of active learning, the authors use the min-margin and max-entropy metrics to compute the informativeness score of each sentence. The curriculum is build by choosing the examples with the best score, according to both difficulty and informativeness, at each step.

\subsection{Speech processing}

\begin{sloppypar}
The collection of articles gathered here show that curriculum learning can also be successfully applied in speech processing tasks. Still, there are less articles trying this approach, when compared to computer vision or natural language processing. One of the reasons might be that an automatic complexity measure for audio data is more difficult to identify.
\end{sloppypar}

\vspace{0.2cm}
\noindent
{\bf Speech recognition.}
Shi et al.~\citeyearpar{shi2015recurrent} address the task of adapting recurrent neural network language models to specific subdomains using curriculum learning. They adapt three curriculum strategies to guide the training from general to (subdomain) specific: Start-from-Vocabulary, Data Sorting, All-then-Specific. Although their approach leads to superior results when the curricula are adapted to a certain scenario, the authors note that the actual data distribution is essential for choosing the right curriculum schedule.

A curriculum approach for speech recognition is illustrated by Amodei et al.~\citeyearpar{amodei2016deep}. They use the length of the utterances to rank the samples from easy to hard. The method consists of training a deep model in increasing order of difficulty for one epoch, before returning to the standard random procedure. This can be regarded as a curriculum warm-up technique, which provides higher quality weights as a starting point from which to continue training the network.

Ranjan and Hansen~\citeyearpar{ranjan2017curriculum} apply curriculum learning for speaker identification in noisy conditions. They use a batching strategy, starting with the easiest subset and progressively adding more challenging batches. The CL approach is used in two distinct stages of a state-of-the-art system: at the probabilistic linear discriminant back-end level, and at the i-Vector extractor matrix estimation level.

Lotfian and Busso~\citeyearpar{lotfian2019curriculum} use curriculum learning for speech emotion recognition. They apply an easy-to-hard batching strategy and fine-tune the learning rate for each bin. The difficulty of the examples is estimated in two ways, using either the error of a pre-trained model or the disagreement between human annotators. Samples that are ambiguous for humans should be ambiguous (more difficult) for the automatic model as well. Another important aspect is that not all annotators have the same expertise. To solve this problem, the authors propose using the minimax conditional entropy to jointly estimate the task difficulty and the rater’s reliability.

\begin{sloppypar}
Zheng et al.~\citeyearpar{zheng2019autoencoder} introduce a semi-supervised curriculum learning approach for speaker verification. The multi-stage method starts with the easiest task, training on labeled examples. In the next stage, unlabeled in-domain data are added, which can be seen as a medium-difficulty problem. In the following stages, the training set is enhanced with unlabeled data from other smart speaker models (out of domain) and with text-independent data, triggering keywords and random speech.
\end{sloppypar}

Caubri{\`e}re et al.~\citeyearpar{caubriere2019curriculum} employ a transfer learning approach based on curriculum learning for solving the spoken language understanding task. The method is multistage, with the data being ordered from the most semantically generic to the most specific. The knowledge acquired at each stage, after each task, is transferred to the following one until the final results are produced.

Zhang et al.~\citeyearpar{zhang2019automatic} propose a teacher-student curriculum approach for digital modulation classification. In the first step, the mentor network is trained using the feedback (loss) from the pre-initialized student network. Then, the student network is trained under the supervision of the curriculum established by the teacher. As the authors argue, this procedure has the advantage of preventing overfitting for the student network. 

Ristea and Ionescu~\citeyearpar{ristea21-interspeech} introduce a self-paced ensemble learning scheme, in which multiple models learn from each other over several iterations. At each iteration, the most confident samples from the target domain and the corresponding pseudo-labels are added to the training set. In this way, an individual model has the chance of learning from the highly-confident labels assigned by another model, thus improving the whole ensemble. The proposed approach shows performance improvements over several speech recognition tasks.

Braun et al.~\citeyearpar{braun2017curriculum} use anti-curriculum learning for automatic speech recognition systems under noisy environments. They use the signal-to-noise ratio (SNR) to create the hard-to-easy curriculum, starting the learning process with low SNR levels and gradually increasing the SNR range to encompass higher SNR levels, thus simulating a batching strategy. The authors also experiment with the opposite ranking of the examples from high SNR to low SNR, but the initial method which emphasizes noisier samples provides better results. 

\vspace{0.2cm}
\noindent
{\bf Other tasks.}
Hu et al.~\citeyearpar{hu2020curriculum} use a curriculum methodology for audiovisual learning. In order to estimate the difficulty of the examples, they build an algorithm to predict the number of sound sources in a given scene. Then, the samples are grouped into batches according to the number of sound-sources and the training commences with the first bin. The easy-to-hard ordering comes from the fact that, in a scene with fewer sound-sources, it is easier to visually localize the sound-makers from the background and align them to the sounds.

Huang et al.~\citeyearpar{huang2020dance} address the task of synthesizing dance movements from music in a curriculum fashion. They use a sequence-to-sequence architecture to process long sequences of music features and capture the correlation between music and dance. The training process starts from a fully guided scheme that only uses ground-truth movements, proceeding with a less guided autoregressive scheme in which generated movements are gradually added. Using this curriculum, the error accumulation of autoregressive predictions at inference is limited. 

Zhang et al.~\citeyearpar{zhang2021enhancing} propose a two-stage curriculum learning approach for improving the performance of audio-visual representations learning. Their teacher-student framework based on constrastive learning starts by pre-training the teacher and then jointly training the teacher and the student models, in the first stage. In the second stage, the roles are reversed, with only the student being trained at first, until commencing the training of both networks.

Wang et al.~\citeyearpar{wang-etal-2020-curriculum} introduce a curriculum pre-training method for speech translation. They claim that the traditional pre-training of the encoder using speech recognition does not provide enough information for the model to perform well. To alleviate this problem, the authors include in their curriculum pre-training approach a basic course for transcription learning and two more complex courses for utterance understanding and word mapping in two languages. Different from other curriculum methods, they do not rank examples from easy to hard, but design a series of tasks with increased difficulty in order to maximize the learning potential of the encoder.

\subsection{Medical imaging}

\begin{sloppypar}
A handful of works show the efficiency of curriculum learning approaches in medical imaging. Although vision-inspired measures, like the input size, should perform well \citep{jesson2017cased}, many of the articles propose a handcrafted curriculum or an order based on human annotators \citep{lotter2017multi,jimenez2019medical,oksuz2019automatic,wei2020learn}.
\end{sloppypar}

\vspace{0.2cm}
\noindent
{\bf Cancer detection and segmentation.}
A two-step curriculum learning strategy is introduced by Lotter et al.~\citeyearpar{lotter2017multi} for detecting breast cancer. In the first step, they train multiple classifiers on segmentation masks of lesions in mammograms. This can be seen as the easy component of the curriculum procedure since the segmentation masks provide a smaller and more precise localization of the lesions. In the second, more difficult stage, the authors use the previously learned features to train the model on the whole image.

Jesson et al.~\citeyearpar{jesson2017cased} introduce a curriculum combined with hard negative mining (HNM) for segmentation or detection of lung nodules on data sets with extreme class imbalance. The difficulty is expressed by the size of the input, with the model initially learning how to distinguish nodules from their immediate surroundings, then gradually adding more global context. Since the vast majority of voxels in typical lung images are non-nodule, a traditional random sampling would show examples with a small effect on the loss optimization. To address this problem, the authors introduce a sampling strategy that favors training examples for which the recent model produces false results.

\vspace{0.2cm}
\noindent
{\bf Other tasks.}
Tang et al.~\citeyearpar{tang2018attention} introduce an attention-guided curriculum learning framework to solve the task of joint classification and weakly-supervised localization of thoracic diseases from chest X-rays. The level of disease severity defines the difficulty of the examples, with training commencing on the more severe samples, and continuing with moderate and mild examples. In addition to the severity of the samples, the authors use the classification probability scores of the current CNN classifier to guide the training to the more confident examples.

Jim{\'e}nez-S{\'a}nchez et al.~\citeyearpar{jimenez2019medical} introduce an easy-to-hard curriculum learning approach for the classification of proximal femur fracture from X-ray images. They design two curriculum methods based on the class difficulty as labeled by expert annotators. The first methodology assumes that categories are equally spaced and uses the rank of each class to assign easiness weights. The second approach uses the agreement of expert human annotators in order to assign the sampling probability. Experiments show the superiority of the curriculum method over multiple baselines, including anti-curriculum designs.

Oksuz et al.~\citeyearpar{oksuz2019automatic} employ a curriculum method for automatically detecting the presence of motion-related artifacts in cardiac magnetic resonance images. They use an easy-to-hard curriculum batching strategy which compares favorably to the standard random approach and to an anti-curriculum methodology. The experiments are conducted by introducing synthetic artifacts with different corruption levels facilitating the easy-to-hard scheduling, from a high to a low corruption level.

Wei et al.~\citeyearpar{wei2020learn} propose a curriculum learning approach for histopathological image classification. In order to determine the curriculum schedule, they use the levels of agreement between seven human annotators. Then, they employ a standard batching approach, splitting the training set into four levels of difficulty and gradually enhancing the training set with more difficult batches. To show the efficiency of the method, they conduct multiple experiments, comparing their results with an anti-curriculum methodology and with different selection criteria.

Alsharid et al.~\citeyearpar{alsharid2020curriculum} employ a curriculum learning approach for training a fetal ultrasound image captioning model. They propose a dual-curriculum approach that relies on a curriculum over both image and text information for the ultrasound image captioning problem. Experimental results show that the best distance metrics for building the curriculum were, in their case, the Wasserstein distance for image data and the TF-IDF metric for text data.

Liu et al.~\citeyearpar{liu-etal-2021-competence} use curriculum learning for solving the medical report generation task. Their apply a two-step approach over which they iterate until convergence. In the first step, they estimate the difficulty of the
training examples and evaluate the competence of the model. Then, they select the appropriate training samples considering the model competence, following the easy-to-hard strategy. To ensure the curriculum schedule, the authors define heuristic and model-based metrics which capture visual and textual difficulty.

Zhao et al.~\citeyearpar{zhao2020egdcl} introduce a curriculum learning approach for improving the computer-aided diagnosis (CAD) of glaucoma. As the authors claim, CAD applications are limited by the data bias, induced by the large number of healthy cases and the hard abnormal cases. To eliminate the bias, the algorithm trains the model from easy to hard and from normal to abnormal. The architecture is a teacher-student framework in which the student provides prior knowledge by identifying the bias of the decision procedure, while the teacher learns the CAD model by resampling the data distribution using the generated curriculum.

Burduja and Ionescu~\citeyearpar{Burduja-ICIP-2021} study model-level and data-level curriculum strategies in medical image alignment. They compare two existing approaches introduced in \citep{Morerio-ICCV-2017,sinha2020curriculum} with a novel approach based on gradually deblurring the input. The latter strategy relies on the intuition that blurred images are easier to align. Hence, the unsupervised training starts with blurred images, which are gradually deblurred until they become identical to the original samples. The empirical results show that curriculum by input blur attains performance gains on par with curriculum by smoothing \citep{sinha2020curriculum}, while reducing the computational complexity by a significant margin.

\subsection{Reinforcement learning}

A large part of the curriculum learning literature focuses on its application in reinforcement learning (RL) settings, usually addressing robotics tasks. Behind this statement stands the extensive survey of Narvekar et al.~\citeyearpar{narkevarJMLR20survey}, which explores this direction in depth. Compared to the curriculum methodologies applied in other domains, most of the approaches used in RL apply the curriculum at task-level, not at data-level. Furthermore, teacher-student frameworks are more common in RL than in the other domains \citep{matiisen2019teacher,narvekar2018learning,portelas2020teacher,portelas2020meta}.

\vspace{0.2cm}
\noindent
{\bf Navigation and control.}
Florensa et al.~\citeyearpar{florensa2017reverse} propose a curriculum approach for reinforcement learning of robotic tasks. The authors claim that this is a difficult problem because the natural reward function is sparse. Thus, in order to reach the goal and receive learning signals, extensive exploration is required. The easy-to-hard methodology is obtained by training the robot in ``reverse'', gradually learning to reach the goal from a set of start states increasingly farther away from the goal. As the distance from the goal increases, so does the difficulty of the task. The nearby, easy, start states are generated from a certain seed state by applying noise in action space.

\begin{sloppypar}
Murali et al.~\citeyearpar{murali2018cassl} also adapt curriculum learning for robotics, learning how to grasp using a multi-fingered gripper. They use curriculum learning in the control space, which guides the training in the control parameter space by fixing some dimensions and sampling in the other dimensions. The authors employ variance-based sensitivity analysis to determine the easy-to-learn modalities that are learned in the earlier phases of the training while focusing on harder modalities later.
\end{sloppypar}

Fournier et al.~\citeyearpar{fournier2019clic} examine a non-rewarding reinforcement learning setting, containing multiple possibly related objects with different values of controllability, where an apt agent acts independently, with non-observable intentions. The proposed framework learns to control individual objects and imitates the agent's interactions. The objects of interest are selected during training by maximizing the learning progress. A sampling probability is computed considering the agent's competence, defined as the average success over a window of tentative episodes at controlling an object, at a certain step.

Fang et al.~\citeyearpar{fang2019curriculum} introduce the Goal-and-Curiosity-driven curriculum learning methodology for RL. Their approach controls the mechanism for selecting hindsight experiences for replay by taking into consideration goal-proximity and diversity-based curiosity. The goal-proximity represents how close the achieved goals are to the desired goals, while the diversity-based curiosity captures how diverse the achieved goals are in the environment. The curriculum algorithm selects a subset of achieved goals with regard to both proximity and curiosity, emphasizing curiosity in the early phases of the training, then gradually increasing the importance of proximity during later episodes. 

\begin{sloppypar}
Manela and Biess~\citeyearpar{MANELA2022260} use a curriculum learning strategy with hindsight experience replay (HER) for solving sequential object manipulation tasks. They train the reinforcement learning agent on gradually more complex tasks in order to alleviate the problem of traditional HER techniques which fail in difficult object manipulation tasks. The curriculum is given by the natural order of the tasks, with all source tasks having the same action spaces. The increase of the state space along the sequence of source tasks captures the easy-to-hard learning strategy very well.
\end{sloppypar}

Luo et al.~\citeyearpar{luo2020accelerating} employ a precision-based continuous CL approach for improving the performance of multi-goal reaching tasks. It consists of gradually adjusting the requirements during the training process, instead of building a static schedule. To design the curriculum, the authors use the required precision as a continuous parameter introduced in the learning process. They start from a loose reach accuracy in order to allow the model to acquire the basic skills required to realize the final task. Then, the required precision is gradually updated using a continuous function. 

Eppe et al.~\citeyearpar{eppe2019curriculum} introduce a curriculum strategy for RL that uses goal masking as a method to estimate a goal's difficulty level. They create goals of appropriate difficulty by masking combinations of sub-goals and by associating a difficulty level to each mask. A training rollout is considered successful if the non-masked sub-goals are achieved. This mechanism allows estimating the difficulty of a previously unseen masked goal, taking into consideration the past success rate of the learner for goals to which the same mask has been applied. Their results suggest that focusing on the medium-difficulty goals is the optimal choice for deep deterministic policy gradient methods, while a strategy where difficult goals are sampled more often produces the best results when hindsight experience replay is employed.

Milano and Nolfi~\citeyearpar{milano2021automated} apply curriculum learning over the evolutionary training of embodied agents. They generate the curriculum by automatically selecting the optimal environmental conditions for the current model. The complexity of the environmental conditions can be estimated by taking into consideration how the agents perform in the chosen conditions. The results on two continuous control optimization benchmarks show the superiority of the curriculum approach.

Tidd et al.~\citeyearpar{tidd2020guided} present a curriculum approach for training deep RL policies for bipedal walking over various challenging terrains. They design the easy-to-hard curriculum using a three-stage framework, gradually increasing the difficulty at each step. The agent starts learning on easy terrain which is gradually enhanced, becoming more complex. In the first stage, the target policy produces forces that are applied to the joints and the base of the robot. These guiding forces are then gradually reduced in the next step, then, in the final step, random perturbations with increasing magnitude are applied to the robot's base to improve the robustness of the policies.

He et al.~\citeyearpar{he2020automatic} introduce a two-level automatic curriculum learning framework for reinforcement learning, composed of a high-level policy, the curriculum generator, and a low-level policy, the action policy. The two policies are trained simultaneously and independently, with the curriculum generator proposing a moderately difficult curriculum for the action policy to learn. By solving the intermediate goals proposed by the high-level policy, the action policy will successfully work on all tasks by the end of the training, without any supervision from the curriculum generator.

Mattisen et al.~\citeyearpar{matiisen2019teacher} introduce the teacher-student curriculum learning (TSCL) framework for reinforcement learning. In this setting, the student tries to learn a complex task, while the teacher automatically selects sub-tasks for the student to learn in order to maximize the learning progress. To address forgetting, the teacher network also chooses tasks where the performance of the student is degrading. Starting from the intuition that the student might not have any success in the final task, the authors choose to maximize the sum of performances in all tasks. As the final task includes elements from all previous tasks, good performance in the intermediate tasks should lead to good performance in the final task. The framework was tested on the addition of decimal numbers with LSTM and navigation in Minecraft.

Portelas et al.~\citeyearpar{portelas2020teacher} also employ a teacher-student algorithm for deep reinforcement learning in which the teacher must supervise the training of the student and generate the right curriculum for it to follow. As the authors argue, the main challenge of this approach comes from the fact that the teacher does not have an initial knowledge about the student's aptitude. To determine the right policy, the problem is translated into a surrogate continuous bandit problem, with the teacher selecting the environments which maximize the learning progress of the student. Here, the authors model the absolute learning progress using Gaussian mixture models. 

Klink et al.~\citeyearpar{pmlr-v100-klink20a} propose a self-paced approach for RL where the curriculum focuses on intermediate distributions and easy tasks first, then proceeds towards the target distribution. The model uses a trade-off between maximizing the local rewards and the global progress of the final task. It employs a bootstrapping technique to improve the results on the target distribution by taking into consideration the optimal policy from previous iterations.

Zhang et al.~\citeyearpar{zhang2020automatic} introduce a curriculum approach for RL focusing on goals of medium difficulty. The intuition behind the technique comes from the fact that goals at the frontier of the set of goals that an agent can reach may provide a stronger learning signal than randomly sampled goals. They employ the Value Disagreement Sampling method, in which the goals are sampled according to the distribution induced by the epistemic uncertainty of the value function. They compute the epistemic uncertainty using the disagreement between an ensemble of value functions, thus obtaining the goals which are neither too hard, nor too easy for the agent to solve.

\vspace{0.2cm}
\noindent
{\bf Games.}
Narvekar et al.~\citeyearpar{narvekar2016source} introduce curriculum learning in a reinforcement learning (RL) setup. They claim that one task can be solved more efficiently by first training the model in a curriculum fashion on a series of optimally chosen sub-tasks. In their setting, the agent has to learn an optimal policy that maximizes the long-term expected sum of discounted rewards for the target task. To quantify the benefit of the transfer, the authors consider asymptotic performance, comparing the final performance of learners in the target task when using transfer with a no transfer approach. The authors also consider a jump-start metric, measuring the initial performance improvement on the target task after the transfer.

Ren et al.~\citeyearpar{ren2018self} propose a self-paced methodology for reinforcement learning. Their approach selects the transitions in a standard curriculum fashion, from easy to hard. They design two criteria for developing the right policy, namely a self-paced prioritized criterion and a coverage penalty criterion. In this way, the framework guarantees both sample efficiency and diversity. The SPL criterion is computed with respect to the relationship between the temporal-difference error and the curriculum factor, while the coverage penalty criterion reduces the sampling frequency of transitions that have already been selected too many times. To prove the efficiency of their method, the authors test the approach on Atari 2600 games.

\vspace{0.2cm}
\noindent
{\bf Other tasks.}
Foglino et al.~\citeyearpar{foglino2019gray} introduce a gray box reformulation of curriculum learning in the RL setup by splitting the task into a scheduling problem and a parameter optimization problem. For the scheduling problem, they take into consideration the regret function, which is computed based on the expected total reward for the final task and on how fast it is achieved. Starting from this, the authors model the effect of learning a task after another, capturing the utility and penalty of each such policy. Using this reformulation, the task of minimizing the regret (thus, finding the optimal curriculum) becomes a parameter optimization problem.

\begin{sloppypar}
Bassich and Kudenko~\citeyearpar{bassich2019continuous} suggest a continuous version of the curriculum for reinforcement learning. For this, they define a continuous decay function, which controls how the difficulty of the environment changes during training, adjusting the environment. They experiment with fixed predefined decays and adaptive decays which take into consideration the performance of the agent. The adaptive friction-based decay, which uses the model from physics with a body sliding on a plane with friction between them to determine the decay, achieves the best results. Experiments also show that higher granularity, i.e.,~a higher frequency for updating the difficulty of an environment during the curriculum, provides better results.
\end{sloppypar}

Nabli and Carvalho~\citeyearpar{nabli2020curriculum} introduce a curriculum-based RL approach to multi-level budgeted combinatorial problems. Their main idea is that, for an agent that can correctly estimate instances with budgets up to $b$, the instances with budget $b + 1$ can be estimated in polynomial time. They gradually train the agent on heuristically solved instances with larger budgets.

Qu et al.~\citeyearpar{qu2018curriculum} use a curriculum-based reinforcement learning approach for learning node representations for heterogeneous star networks. They suggest that the learning order of different types of edges significantly impacts the overall performance. As in the other RL applications, the goal is to find the policy that maximizes the cumulative rewards. Here, the reward is computed as the performance on external tasks, where node representations are considered as features. 

Narvekar and Stone~\citeyearpar{narvekar2018learning} extend previous curriculum methods for reinforcement learning that formulate the curriculum sequencing problem as a Markov Decision Process to multiple transfer learning algorithms. Furthermore, they prove that curriculum policies can be learned. In order to find the state in which the target task is solved in the least amount of time, they represent the state as a set of potential functions which take into consideration the previously sampled source tasks.

Turchetta et al.~\citeyearpar{turchetta2020safe} present an approach for identifying the optimal curriculum in safety-critical applications where mistakes can be very costly. They claim that, in these settings, the agent must behave safely not only after but also during learning. In their algorithm, the teacher has a set of reset controllers which activate when the agent starts behaving dangerously. The set takes into consideration the learning progress of the students in order to determine the right policy for choosing the reset controllers, thus optimizing the final reward of the agent.

\begin{sloppypar}
Portelas et al.~\citeyearpar{portelas2020meta} introduce the idea of meta automatic curriculum learning for RL, in which the models are learning ``to learn to teach''. Using knowledge from curricula built for previous students, the algorithm improves the curriculum generation for new tasks. Their method combines inferred progress niches with the learning progress based on the curriculum learning algorithm from \citep{portelas2020teacher}. In this way, the model adapts towards the characteristics of the new student and becomes independent of the expert teacher once the trajectory is completed. 
\end{sloppypar}

Feng et al.~\citeyearpar{feng2020novel} propose a curriculum-based RL approach in which, at each step of the training, a batch of task instances are fed to the agent which tries to solve them, then, the weights are adjusted according to the results obtained by the agent. The selection criterion differs from other methods, by not choosing the easiest tasks, but the tasks which are at the limit of solvability.

\subsection{Other domains}

Aside from the previously presented works, there are few papers which do not fit in any of the explored domains. 
For example, Zhao et al.~\citeyearpar{zhao2015self} propose a self-paced easy-to-complex approach for matrix factorization. Similar to previous methods, they build a regularizer which, based on the current loss, favors easy examples in the first rounds of the training. Then, as the model ages, it gives the same weight to more difficult samples. Different from other methods, they do not use a hard selection of samples, with binary (easy or hard) labels. Instead, the authors propose a soft approach, using real numbers as difficulty weights, in order to faithfully capture the importance of each example.

Graves et al.~\citeyearpar{graves2016hybrid} introduce a differentiable neural computer, a model consisting of a neural network that can perform read-write operations to an external memory matrix. In their graph traversal experiments, they employ an easy-to-hard curriculum method, where the difficulty is calculated using task-dependent metrics (i.e., the number of nodes in the graph). They build the curriculum using a linear sequence of lessons in ascending order of complexity.

Ma et al.~\citeyearpar{ma2018convergence} conduct an extensive theoretical analysis with convergence results of the implicit SPL objective. By proving that the SPL process converges to critical points of the implicit objective when used in light conditions, the authors verify the intrinsic relationship between self-paced learning and the implicit objective. These results prove that the robustness analysis on SPL is complete and theoretically sound. 

Zheng et al.~\citeyearpar{zheng2020unsupervised} introduce the self-paced learning regularization to the unsupervised feature selection task. The traditional unsupervised feature selection methods remove redundant features but do not eliminate outliers. To address this issue, the authors enhance the method with an SPL regularizer. Since the outliers are not evenly distributed across samples, they employ an easy-to-hard soft weighting approach over the traditional hard threshold weight.

Sun and Zhou~\citeyearpar{sun2020fspmtl} introduce a self-paced learning method for multi-task learning which starts training on the simplest samples and tasks, while gradually adding the more difficult examples. In the first step, the model obtains sample difficulty levels to select samples from each task. After that, samples of different difficulty levels are selected, taking a standard SPL approach that uses the value of the loss function. Then, a high-quality model is employed to learn data iteratively until obtaining the final model. The authors claim that using this methodology solves the scalability issues of other approaches, in limited data scenarios. 

\begin{sloppypar}
Zhang et al.~\citeyearpar{zhang2020worst} propose a new family of worst-case-aware losses across tasks for inducing automatic curriculum learning in the multi-task setting. Their model is similar to the framework of Graves et al.~\citeyearpar{Graves-ICML-2017} and uses a multi-armed bandit with an arm for each task in order to learn a curriculum in an online fashion. The training examples are selected by choosing the task with the likelihood proportional to the average loss or the task with the highest loss. Their worst-case-aware approach to generate the policy provides good results for zero-shot and few-shot applications in multi-task learning settings. 
\end{sloppypar}

\begin{figure*}
\begin{center}
\includegraphics[width=0.99\linewidth]{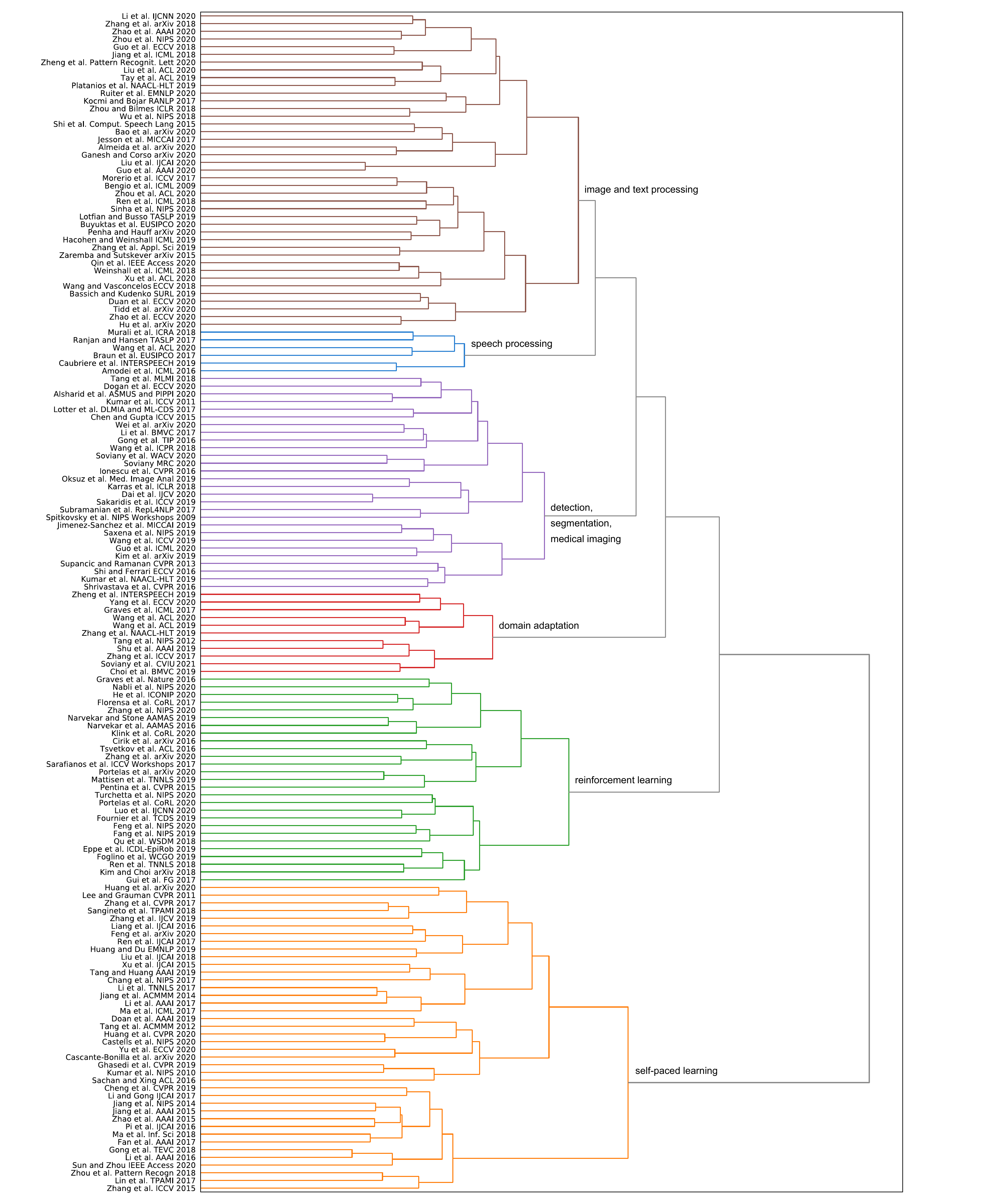}
\end{center}
\vspace*{-0.6cm}
\caption{Dendrogram of curriculum learning articles obtained using agglomerative clustering. Best viewed in color.}
\label{fig_clustering}
\vspace*{-0.4cm}
\end{figure*}

\section{Hierarchical Clustering of Curriculum Learning Methods}
\label{sec_clustering}

Since the taxonomy described in the previous section can be biased by our subjective point of view, we also build an automated grouping of the curriculum learning papers. To this end, we employ an agglomerative clustering algorithm based on Ward's linkage. We opt for this hierarchical clustering algorithm because it performs well even when there is noise between clusters. Since we are dealing with a fine-grained clustering problem, i.e.,~all papers are of the same genre (scientific) and on the same topic (namely, curriculum learning), eliminating as much of the noise as possible is important. We represent each scientific article as a term frequency-inverse document frequency (TF-IDF) vector over the vocabulary extracted from all the abstracts. The purpose of the TF-IDF scheme is to reduce the chance of clustering documents based on words expected to be common in our set of documents, such as ``curriculum'' or ``learning''. Although TF-IDF should also reduce the significance of stop words, we decided to eliminate stop words completely. The result of applying the hierarchical clustering algorithm, as described above, is the tree (dendrogram) illustrated in Figure~\ref{fig_clustering}.

By analyzing the resulting dendrogram, we observed a set of homogeneous clusters, containing at most one or two outlier papers. The largest homogeneous cluster (orange) is mainly composed of self-paced learning methods \citep{zhang2015self,zhou2018deep,sun2020fspmtl,li2016multi,gong2018decomposition,fan2016self,ma2018convergence,pi2016self,zhao2015self,Jiang-AAAI-2015,jiang2014self,sachan2016easy,Kumar-NIPS-2010,Ghasedi-CVPR-2019,cascante2020curriculum,castells2020superloss,huang2020curricularface,tang2012self,ma2017self,li2016self,jiang2014easy,li2017self-b,chang2017active,tang2019self,xu2015multi,ren2017robust,feng2020semi,liang2016learning,Zhang-IJCV-2019,lee2011learning}, while the second largest homogeneous cluster (green) is formed of reinforcement learning methods \citep{ren2018self,foglino2019gray,eppe2019curriculum,qu2018curriculum,fang2019curriculum,feng2020novel,fournier2019clic,luo2020accelerating,portelas2020teacher,turchetta2020safe,pentina2015curriculum, matiisen2019teacher, portelas2020meta, sarafianos2017curriculum,zhang2020worst,florensa2017reverse,he2020automatic,nabli2020curriculum}. It is interesting to see that the self-paced learning cluster (depicted in orange) is joined with the rest of the curriculum learning methods at the very end, which is consistent with the fact that self-paced learning has developed as an independent field of study, not necessarily tied to curriculum learning. Our dendrogram also indicates that the curriculum learning strategies applied in reinforcement learning (green cluster) is a distinct breed than the curriculum learning strategies applied in other domains (brown, blue, purple and red clusters). Indeed, in reinforcement learning, curriculum strategies are typically based on teacher-student models
or are applied over tasks, while in the other domains, curriculum strategies are commonly applied over data samples. The third largest homogeneous cluster (red) is mostly composed of domain adaptation methods \citep{Choi2019PseudoLabelingCF,soviany2019curriculum,zhang2017curriculum,shu2019transferable,tang2012shifting,zhang2019curriculum,wang-etal-2019-dynamically,wang-etal-2020-curriculum,Graves-ICML-2017,YangBLS20,zheng2019autoencoder}. In the cross-domain setting, curriculum learning is typically designed to gradually adjust the model from the source domain to the target domain. Hence, such curriculum learning methods can be seen as domain adaptation approaches. Finally, our last homogeneous cluster (blue) contains speech processing methods \citep{ranjan2017curriculum,braun2017curriculum,caubriere2019curriculum, amodei2016deep}. We are thus left with two heterogeneous clusters (brown and purple). The largest heterogeneous cluster (brown) is equally dominated by text processing methods \citep{li2020label,zhang2018empirical,ZhaoWNW20,liu-etal-2020-norm,tay-etal-2019-simple,platanios2019competence,ruiter2020self,kocmi2017curriculum,wu2018learning,shi2015recurrent,bao2020plato,liu2020task,guo2020fine,Bengio-ICML-2009,zhou2020uncertainty,penha2019curriculum,zaremba2014learning,xu2020curriculum} and image classification approaches \citep{zhou2020curriculum,guo2018curriculumnet,jiang2018mentornet,zhou2018minimax,wu2018learning,ganesh2020rethinking,Morerio-ICCV-2017,ren18a,sinha2020curriculum,Hacohen2019OnTP,qin2020balanced,Weinshall2018CurriculumLB,wang2018towards}. However, we were not able to identify representative (homogeneous) subclusters for these two domains. The second largest heterogeneous cluster (purple) is dominated by works that study object detection \citep{Chen_2015_ICCV,Li-BMVC-2017,Wang-ICPR-2018,soviany2020curriculum,saxena2019data,shrivastava2016training}, semantic segmentation \citep{dai2020curriculum,sakaridis2019guided} and medical imagining \citep{tang2018attention,lotter2017multi,wei2020learn,oksuz2019automatic,jimenez2019medical}. While the tasks gathered in this cluster are connected at a higher level (being studied in the field of computer vision), we were not able to identify representative subclusters.

In summary, the dendrogram illustrated in Figure~\ref{fig_clustering} suggests that curriculum learning works should be first divided by the underlying learning paradigm: supervised learning, reinforcement learning, and self-paced learning. At the second level, the scientific works that fall in the cluster of supervised learning methods can be further divided by task or domain of application: image classification and text processing, speech processing, object detection and segmentation, and domain adaptation. We thus note our manually determined taxonomy is consistent with the automatically computed hierarchical clustering.

\section{Closing Remarks and Future Directions}
\label{sec_conclusion}

\subsection{Generic directions}

\begin{sloppypar}
\noindent
{\bf Curriculum learning may degrade data diversity and produce worse results.}
While exploring the curriculum learning literature, we observed that curriculum learning was successfully applied in various domains, including computer vision, natural language processing, speech processing and robotic interaction. Curriculum learning has brought improved performance levels in tasks ranging from image classification, object detection and semantic segmentation to neural machine translation, question answering and speech recognition. However, we note that curriculum learning is not always bringing significant performance improvements. We believe this happens because there are other factors that influence performance, and these factors can be negatively impacted by curriculum learning strategies. For example, if the difficulty measure has a preference towards choosing easy examples from a small subset of classes, the diversity of the data samples is affected in the preliminary training stages. If this problem occurs, it could lead to a suboptimal training process, guiding the model to a suboptimal solution. This example shows that, while employing a curriculum learning strategy, there are other factors that can play key roles in achieving optimal results. We believe that exploring the side effects of curriculum learning and finding explanations for the failure cases is an interesting line of research for the future. Studies in this direction might lead to a generic successful recipe for employing curriculum learning, subject to the possibility of controlling the additional factors while performing curriculum learning.
\end{sloppypar}

\noindent
{\bf Model-level and performance-level curriculum is not sufficiently explored.}
Regarding the components implied in Definition~\ref{def_ML}, we noticed that the majority of curriculum learning approaches perform curriculum with respect to the experience E, this being the most natural way to apply curriculum. Another large body of works, especially those on reinforcement learning, studied curriculum with respect to the class of tasks T. The success of such curriculum learning approaches is strongly correlated with the characteristics of the difficulty measure used to determine which data samples or tasks are easy and which are hard. Indeed, a robust measure of difficulty, for example the one that incorporates diversity \citep{soviany2020curriculum}, seems to bring higher improvements compared to a measure that overlooks data sample diversity. However, we should emphasize that the other types of curriculum learning, namely those applied on the model M or the performance measure P, do not necessarily require an explicit formulation of a difficulty measure. Contrary to our expectation, there seems to be a shortage of such studies in literature. A promising and generic approach in this direction was recently proposed by Sinha et al.~\citeyearpar{sinha2020curriculum}. However, this approach, which performs curriculum by deblurring convolutional activation maps to increase the capacity of the model, studies mainstream vision tasks and models. In future work, curriculum by increasing the learning capacity of the model can be explored by investigating more efficient approaches and a wider range of tasks.

\begin{sloppypar}
\noindent
{\bf Curriculum is not sufficiently explored in unsupervised and self-supervised learning.}
Curriculum learning strategies have been investigated in conjunction with various learning paradigms, such as supervised learning, cross-domain adaptation, self-paced learning, semi-supervised learning and reinforcement learning. Our survey uncovered a deficit of curriculum learning studies in the area of unsupervised learning and, more specifically, self-supervised learning. Self-supervised learning is a recent and hot topic in domains such as computer vision \citep{Wei-CVPR-2018} and natural language processing \citep{Devlin-NAACL-2019}, that developed, in most part, independently of the body of curriculum learning works. We believe that curriculum learning may play a very important role in unsupervised and self-supervised learning. Without access to labels, learning from a subset of easy samples may offer a good starting point for the optimization of an unsupervised model. In this context, a less diverse subset of samples, at least in the preliminary training stages, could prove beneficial, contrary to the results shown in supervised learning tasks. In self-supervision, there are many approaches, e.g., Georgescu et al.~\citeyearpar{georgescu2020anomaly}, showing that multi-task learning is beneficial. Nonetheless, the order in which the tasks are learned might influence the final performance of the model. Hence, we consider that a significant amount of attention should be dedicated to the development of curriculum learning strategies for unsupervised and self-supervised models.
\end{sloppypar}

\noindent
{\bf The connection between curriculum learning and SGD is not sufficiently understood.}
We should emphasize that curriculum learning is an approach that is typically applied on neural networks, since changing the order of the data samples can influence the performance of such models. This is tightly coupled with the fact that neural networks have non-convex objectives. The mainstream approach to optimize non-convex models is based on some variation of stochastic gradient descent (SGD). The fact that the order of data samples influences performance is caused by the stochasticity of the training process. This observation exposes the link between SGD and curriculum learning, which might not be obvious at the first sight. On the positive side, SGD enables the possibility to apply curriculum learning on neural models in a straightforward manner. Since curriculum learning usually implies restricting the set of samples to a subset of easy samples in the preliminary training stages, it might constrain SGD to converge to a local minimum from which it is hard to escape as increasingly difficult samples are gradually added. Thus, the negative side is that curriculum learning makes it harder to control the optimization process, requiring additional babysitting. We believe this is the main factor that leads to convergence failures and inferior results when curriculum learning is not carefully integrated in the training process. One potential direction of future research is proposing solutions that can automatically regulate the curriculum training process. Perhaps an even more promising direction is to couple curriculum learning strategies with alternative optimization algorithms, e.g.~evolutionary search. We can go as far as saying that curriculum learning could even close the gap between the widely used SGD and other optimization methods.

\subsection{Domain-specific directions}

\noindent
{\bf Curriculum learning in computer vision.}
The current focus of the computer vision researchers is the development of vision transformers \citep{Carion-ECCV-2020,Caron-ICCV-2021,Dosovitskiy-ICLR-2021,Jaegle-ICML-2021,Khan-arXiv-2021,Wu-arXiv-2021,Zhu-ICLR-2020}, which make use of the global information and self-repeating patterns to reach record-high performance levels across a broad range of tasks. Transformers are usually trained in two stages, namely a pre-training stage on large scale data using self-supervision, and a fine-tuning stage on downstream tasks using classic supervision. In this context, we believe that curriculum learning can be employed in either training stage, or even both. In the pre-training stage, it is likely that organizing the self-supervised tasks in the increasing order of complexity would lead to faster convergence, thus being a good topic for future work. In the fine-tuning stage, we would recommend data-level curriculum, e.g. using an image difficulty predictor attentive to data diversity, and model-level curriculum, e.g. gradually unsmoothing tokens, to obtain accuracy and efficiency gains in future research. To our knowledge, curriculum learning has not been applied to vision transformers so far.

\noindent
{\bf Curriculum learning in medical imaging.}
Following the new trend in computer vision, an emerging area of research in medical imaging is about transformers \citep{Chen-arXiv-2021,Chen-VITVN-2021,Gao-arXiv-2021,Hatamizadeh-arXiv-2021,Korkmaz-arXiv-2021,Luthra-arXiv-2021,ristea2021cytran}. Since curriculum learning has not been studied in conjunction with medical image transformers, this seems like a promising direction for future research. However, for the data-level curriculum, we should take into account that difficult images are often those containing both healthy tissue and lesions, while being weakly labeled. Hence, the curriculum could start with images that represent either completely healthy tissue or predominantly lesions, which should be easier to discriminate.

\noindent
{\bf Curriculum learning in natural language processing.}
In natural language processing, language transformers such as BERT \citep{Devlin-NAACL-2019} and GPT-3 \citep{Brown-NeurIPS-2020} have become widely adopted, representing the new norm when it comes to language modeling. Some researchers \citep{xu2020curriculum,Zhang-ISPASS-2021} have already tried to apply curriculum learning strategies to improve language transformers. However, in many cases, the curriculum is based on simple heuristics, such as text length \citep{tay-etal-2019-simple,Zhang-ISPASS-2021}. However, a short text is not always easier to comprehend. We conjecture that a promising direction is to design a curriculum that better resembles our own human learner experience. When humans learn to speak or write in a native or foreign language, they start with a limited vocabulary that progressively expands. Thus, the natural way to perform curriculum should simply be based on the size of the vocabulary. In pursuing this direction, we will need to determine what words should be included in the initial vocabulary and when to expand the vocabulary.

\noindent
{\bf Curriculum learning in signal processing.}
While the machine learning methods employed in signal processing have similar architectural designs to methods employed in computer vision or other fields, the technical challenges are different. Some of the domain-specific challenges are related to signal denoising and source separation. To solve such challenges, we could design specific curriculum learning strategies in future work, e.g. organize the data samples according to the noise level or the number of sources.

\begin{acknowledgements}
The authors would like to thank the reviewers for their useful feedback.
\end{acknowledgements}

%
\section*{Conflict of interest}
The authors declare that they have no conflict of interest.

\bibliographystyle{spbasic}      
\bibliography{biblio}   

\end{document}